\documentclass[final,3p,times]{elsarticle}

\usepackage{amssymb}
\usepackage{amsmath}

\usepackage{lineno}

\journal{Applied Soft Computing}

\usepackage{booktabs}
\usepackage{longtable}
\usepackage{url}
\usepackage{algorithm}
\usepackage{algorithmic}
\usepackage[colorlinks]{hyperref}
\usepackage{multirow}
\usepackage{graphicx}
\usepackage{amsmath}
\usepackage{subcaption}
\usepackage[dvipsnames,table]{xcolor}
\usepackage{bbm}
\usepackage{bm}
\usepackage{coffeestains}

\usepackage{lscape}
\usepackage{array}
\usepackage{setspace}

\newdefinition{definition}{Definition}

\begin{document}

\begin{frontmatter}



\title{Crafting Imperceptible On-Manifold Adversarial Attacks for Tabular Data} 


\author[inst1]{Zhipeng He\corref{cor1}}
\ead{zhipeng.he@hdr.qut.edu.au}
\author[inst3]{Alexander Stevens}
\ead{alexander.stevens@kuleuven.be}
\author[inst1]{Chun Ouyang}
\ead{c.ouyang@qut.edu.au}
\author[inst3]{Johannes De Smedt}
\ead{johannes.desmedt@kuleuven.be}
\author[inst1]{Alistair Barros}
\ead{alistair.barros@qut.edu.au}
\author[inst4,inst5]{Catarina Moreira}
\ead{catarina.pintomoreira@uts.edu.au}

\affiliation[inst1]{organization={School of Information Systems, Queensland University of Technology},
            city={Brisbane},
            country={Australia}}


\affiliation[inst3]{organization={Research Centre for Information Systems Engineering (LIRIS), KU Leuven},
            city={Brussels},
            country={Belgium}}

\affiliation[inst4]{organization={Data Science Institute, University of Technology},
            city={Sydney},
            country={Australia}}

\affiliation[inst5]{organization={INESC-ID/Instituto Superior Técnico, University of Lisboa},
            city={Lisboa},
            country={Portugal}}

\cortext[cor1]{Corresponding author}

\begin{abstract}
Adversarial attacks on tabular data present unique challenges due to the heterogeneous nature of mixed categorical and numerical features. Unlike images where pixel perturbations maintain visual similarity, tabular data lacks intuitive similarity metrics, making it difficult to define \emph{imperceptible} modifications.
Additionally, traditional gradient-based methods prioritise $\ell_p$-norm constraints, often producing adversarial examples that deviate from the original data distributions, making them detectable. To address this, we propose a latent-space perturbation framework using a mixed-input Variational Autoencoder (VAE) to generate statistically consistent adversarial examples. The proposed VAE integrates categorical embeddings and numerical features into a unified latent manifold, enabling perturbations that preserve statistical consistency. We introduce \emph{In-Distribution Success Rate} (IDSR) to jointly evaluate attack effectiveness and distributional alignment. Evaluation across six publicly available datasets and three model architectures  demonstrates that our method achieves substantially lower outlier rates and more consistent performance compared to traditional input-space attacks and other VAE-based methods adapted from image domain approaches, achieving substantially lower outlier rates and higher IDSR across six datasets and three model architectures. Our comprehensive analyses of hyperparameter sensitivity, sparsity control, and generative architecture demonstrate that the effectiveness of VAE-based attacks depends strongly on reconstruction quality and the availability of sufficient training data. When these conditions are met, the proposed framework achieves superior practical utility and stability compared with input-space methods.
This work underscores the importance of maintaining on-manifold perturbations for generating realistic and robust adversarial examples in tabular domains. 

\end{abstract}




\end{frontmatter}


\section{Introduction}

Adversarial attacks involve strategically introducing small, often imperceptible perturbations to input data to deceive machine learning models~\citep{goodfellow2015explaining}. These attacks play critical roles in exposing model vulnerabilities and improving robustness through adversarial training~\citep{szegedy2014intriguing} for both classification~\citep{goodfellow2015explaining,akhtar2021advances} and regression tasks~\citep{guo2023deep,guo2024knowledge}. While extensively studied in continuous domains like images~\citep{akhtar2018threat,akhtar2021advances}, adversarial example generation for tabular data --- a cornerstone of healthcare \citep{allgaier2023does}, finance \citep{ozbayoglu2020deep}, cybersecurity \citep{li2025malware} and Internet of Things \citep{xu2023data,bilal2024blockchain,ahmed2025enhancing} --- remains underexplored despite its real-world significance. 

\begin{figure}[t]
    \centering
    \begin{subfigure}[t]{0.5\columnwidth}
        \centering
        \includegraphics[width=\textwidth]{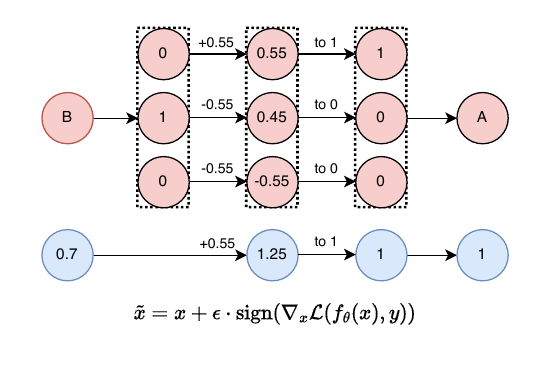}
        \caption{Apply FGSM on both categorical and numerical features.}
        \label{fig:intro_1}
    \end{subfigure}
    \hfill
    \begin{subfigure}[t]{0.48\columnwidth}
        \centering
        \includegraphics[width=\textwidth]{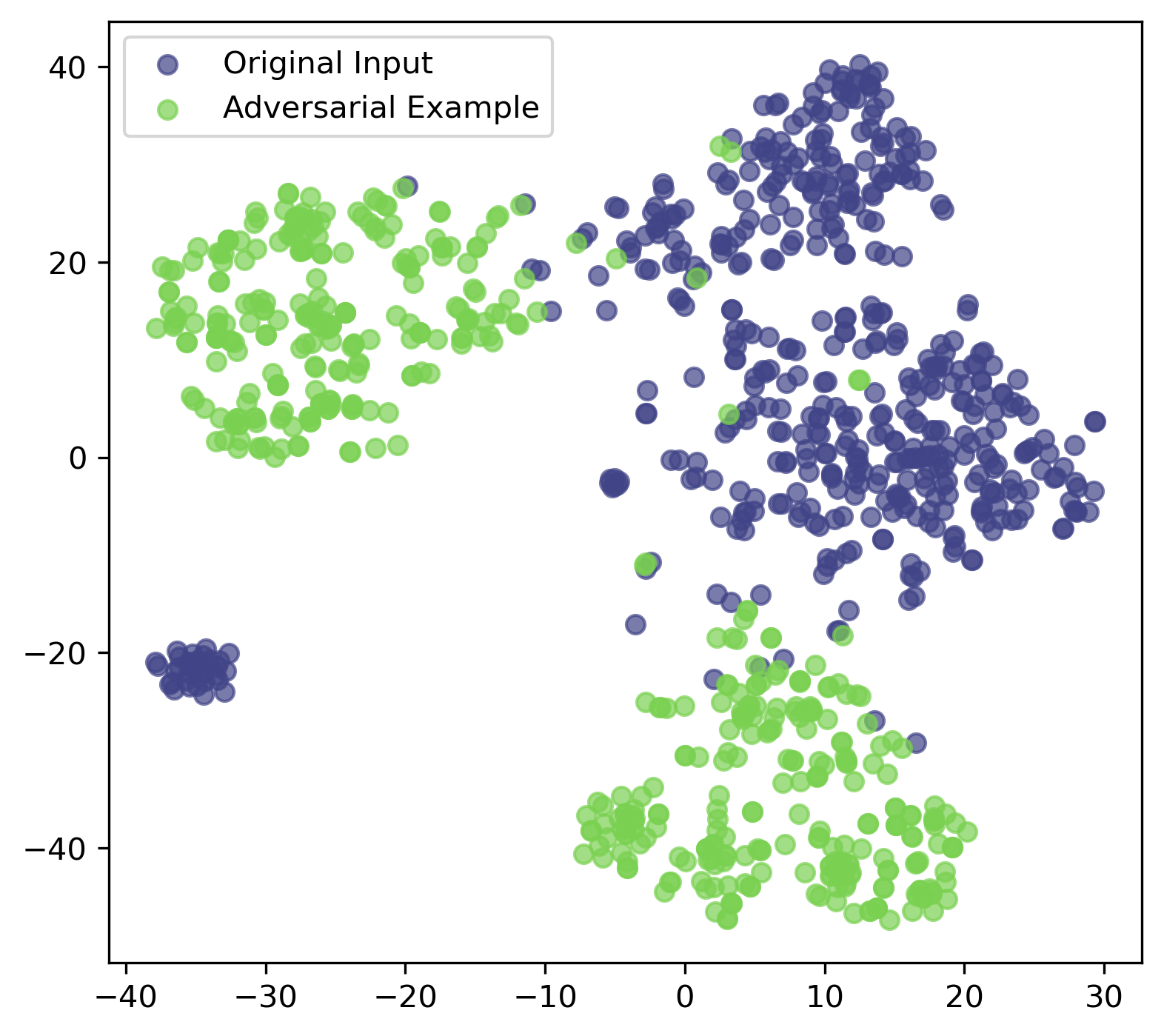}
        \caption{Out-of-distribution adversarial examples by FGSM on Adult.}
        \label{fig:intro_2}
    \end{subfigure}
    \caption{(a) Illustration of applying the Fast Gradient Sign Method (FGSM) to tabular data, demonstrating the perturbation of categorical and numerical features. (b) A scatter plot visualises that the adversarial examples (green) by FGSM deviate from the original input distribution (blue) on Adult dataset.}
    \label{fig:inro}
\end{figure}

Tabular data presents unique challenges compared to homogeneous domains like images. It combines categorical features and numerical features, which require distinct preprocessing. Categorical features are typically one-hot encoded or embedded into numerical representations, inflating dimensionality and sparsity \citep{borisov2022deep}. Numerical features, meanwhile, follow heterogeneous scales and distributions. As illustrated in Figure~\ref{fig:intro_1}, perturbing tabular data involves distinct adjustments for categorical and numerical features.
Adversarial attacks typically use the $\ell_2$ norm to constrain perturbation magnitude and ensure imperceptibility~\citep{tramer2019adversarial}. However, this standard distance metric creates fundamental problems for tabular data: minor adjustments to encoded categorical features (e.g., flipping a one-hot category from [0,1,0] to [1,0,0]) produce large, fixed $\ell_2$ distances, while equivalent numerical perturbations can achieve the same model deception with much smaller $\ell_2$ budgets. This imbalance forces adversarial attack algorithms to either severely restrict categorical modifications or allow disproportionately large perturbations, hindering the generation of effective and realistic adversarial examples.


\subsection{Imperceptibility in Tabular Attack Settings}

The concept of imperceptibility in tabular adversarial attacks can be approached in different ways. \citet{he2025investigating} categorise imperceptibility into seven properties. Among these, \textit{\textbf{deviation}} is critical, as it ensures that adversarial examples remain statistically consistent with the original dataset. This principle requires perturbations to preserve the dataset’s inherent statistical patterns. Violations of these patterns, such as altering distribution shapes, would render adversarial examples detectable as anomalies. 

Popular gradient-based attacks like Fast Gradient Sign Method (FGSM)~\citep{goodfellow2015explaining} and Projected Gradient Descent (PGD)~\citep{madry2017towards} face inherent limitations in tabular domains. While effective in continuous spaces (e.g., images), their reliance on $\ell_p$-norms (e.g., $\|\tilde{x} - x\|_p < \epsilon$)~\citep{tramer2019adversarial} proves inadequate for maintaining the statistical consistency in tabular data. As shown in Figure~\ref{fig:intro_2}, FGSM-generated adversarial examples visibly deviate from the original Adult dataset distribution. This occurs because $\ell_p$-norms optimise for confined continuous space perturbations rather than preserving feature correlations or categorical boundaries, often producing distributional outliers. This highlights the need for perturbation methods tailored to tabular feature spaces, which account for both the preservation of statistical dependencies and the discrete nature of categorical attributes.

Existing adversarial attack methods for tabular data remain constrained by their reliance on direct feature-space perturbations and norm-based distance metrics~\citep{ballet2019imperceptible,simonetto2021unified,kireev2022adversarial,he2025investigating}. Such approaches treat all attributes as continuous and independent, overlooking the dependencies and discrete boundaries that characterise real-world tabular data \citep{borisov2022deep}. Consequently, they often produce perturbations that distort marginal or joint distributions, or require disproportionately large magnitudes to induce misclassification. Moreover, without an explicit generative mechanism, these attacks cannot ensure that perturbed samples remain within the support of the original data distribution. This limitation motivates the development of generative frameworks that maintain distributional validity while enabling effective adversarial manipulation.


\subsection{Latent Space Perturbations}
To overcome the challenges of imperceptibility and deviation in tabular adversarial attacks, we propose generating adversarial examples in the latent space of a Variational Autoencoder (VAE)~\citep{kingma2013auto}, rather than directly manipulating the input feature space. 
Our VAE architecture achieves effective latent representation for tabular data through three mechanisms: (1) learned embeddings transform categorical variables into dense vectors, (2) shared encoder layers unify mixed inputs into a coherent representation and establish cross-feature relationships, and (3) specialised decoders reconstruct features with type-specific losses. This unified latent manifold mitigates encoding biases while preserving feature semantics. This approach provides two critical advantages over direct feature-space perturbations: it eliminates encoding bias by representing all features in a continuous manifold, avoiding the sparsity issues of one-hot encoding~\citep{borisov2022deep}, while ensuring on-manifold consistency through latent-space transformations that maintain statistical indistinguishability from input data. 




\subsection{Contribution}

Our work addresses the challenges of generating imperceptible adversarial examples for tabular data through latent space perturbations. The key contributions are:
\begin{itemize}
    \item We develop a mixed-input VAE integrated with a classification head to process numerical and categorical features, enabling adversarial example generation in a unified continuous latent space.
    \item Our method generates adversarial examples by perturbing latent vectors, ensuring statistical consistency with original data distributions via manifold alignment.
    \item We introduce a novel evaluation metric, \textit{In-Distribution Success Rate} (IDSR), to measure the extent to which tabular adversarial examples deviate from input data distributions.
    \item Through comprehensive evaluation across six datasets and three model architectures, we demonstrate that our approach achieves substantially lower outlier rates and more consistent performance compared to traditional gradient-based attacks (FGSM, PGD) and other VAE-based methods adapted from image domain approaches.
    \item We provide a systematic analysis of performance factors including hyperparameter sensitivity, sparsity control mechanisms, and architectural choices, revealing the critical importance of reconstruction quality for VAE-based adversarial attacks on tabular data.
\end{itemize}

\section{Related Works}

\paragraph{Adversarial Attacks on Tabular Data} 

Recent research on adversarial attacks for tabular data has focused on two key challenges: designing effective attack strategies and ensuring imperceptibility of perturbations. 

\textit{Effective attack strategies} emphasise computational efficiency and highly successful manipulation. \citet{gressel2021feature} introduced model-agnostic attacks guided by feature importance metrics to perturb high-impact attributes, while \citet{chang2023fast} demonstrated efficient label-flipping attacks that compromise model integrity with minimal computational overhead. 

\textit{Imperceptibility} addresses the need for realistic perturbations that preserve data integrity. Early work by \citet{ballet2019imperceptible} proposed LowProFool, which measures perturbation visibility using a weighted $\ell_2$ distance incorporating feature importance. Subsequent studies refined this concept: \citet{nandy2023non} developed gradient-based methods that respect discrete feature constraints and \citet{Chernikova2022FENCE} enforced domain-specific feasibility rules. Constrained adaptive attack~\citep{simonetto2024constrained} advanced this field by combining gradient-based and search-based techniques to generate imperceptible adversarial examples efficiently. They also established a benchmarking framework to further assess the proposed attack across diverse datasets and models \citep{simonetto2024tabularbench}. Finally, \citet{he2025investigating} systematically defined imperceptibility through seven properties, enabling unified quantitative and qualitative assessment of adversarial perturbations, and further introduced a dedicated benchmarking framework that evaluates these properties consistently across tabular datasets and attack methods \citep{he2025tabattackbench}.

In this work, we adopt the deviation property proposed in \citet{he2025investigating} to ensure that adversarial examples remain statistically aligned with the original dataset distribution. 

\paragraph{On-manifold Adversarial Attacks} 

On-manifold adversarial attacks aim to create adversarial examples that remain within the original data distribution. These methods rely on generative models, such as VAEs and GANs, to capture the underlying structure of the data, ensuring that perturbations produce realistic and imperceptible adversarial examples. Previous research has demonstrated the effectiveness of latent space manipulation for adversarial attacks, especially in image-based applications. For instance, \citet{stutz2019disentangling} showed that adversarial robustness and generalisation can coexist through on-manifold perturbations that align with the data distribution. Similarly, \citet{clare2023generating} utilised generative models to apply diverse latent-space perturbations, improving classifier robustness. VAEs have also been effective in generating adversarial examples for multivariate time series~\citep{harford2021generating} and enabling on-manifold attacks in business process monitoring~\citep{stevens2023manifold}.

However, existing VAE-based adversarial generation methods primarily focus on images, time series, and process monitoring, leaving tabular data underexplored. Our work extends VAE-based latent-space perturbation techniques to generate on-manifold adversarial examples for tabular data, ensuring that perturbations adhere to feature correlations and statistical distributions.

\section{On-Manifold Adversarial Attacks for Tabular Data}\label{sec:method}

\paragraph{Problem Definition}

Let $\mathcal{D} = \{(x_i, y_i)\}_{i=1}^N$ denote a tabular dataset where each input $x_i$ is composed of categorical and numerical features: $x_i = \big(x_i^{\text{cat}}, x_i^{\text{num}}\big)$. Here, $x_i^{\text{cat}}$ belongs to the Cartesian product $\mathcal{C} = \mathcal{C}_1 \times \cdots \times \mathcal{C}_m$, representing $m$ categorical attributes, while $x_i^{\text{num}} \in \mathbb{R}^d$ corresponds to a $d$-dimensional vector of numerical features. The label associated with $x_i$ is denoted by $y_i \in \mathcal{Y}$. A classifier $f_\theta: \mathcal{C} \times \mathbb{R}^d \to \mathcal{Y}$, trained on $\mathcal{D}$, maps the concatenated inputs $[x^{\text{cat}}, x^{\text{num}}]$ to predicted labels. To represent the classifier's internal scoring function, we define $Z(x) \in \mathbb{R}^K$ as the \textit{logit} vector (pre-softmax outputs) produced by the same model before applying the final decision rule, such that $f_\theta(x) = \arg\max_i Z(x)_i$. Each element $Z(x)_i$ corresponds to the unnormalised confidence score for class $i$.

\begin{figure*}
    \centering
    \includegraphics[trim={0 0 3.4cm 0},clip,width=\linewidth]{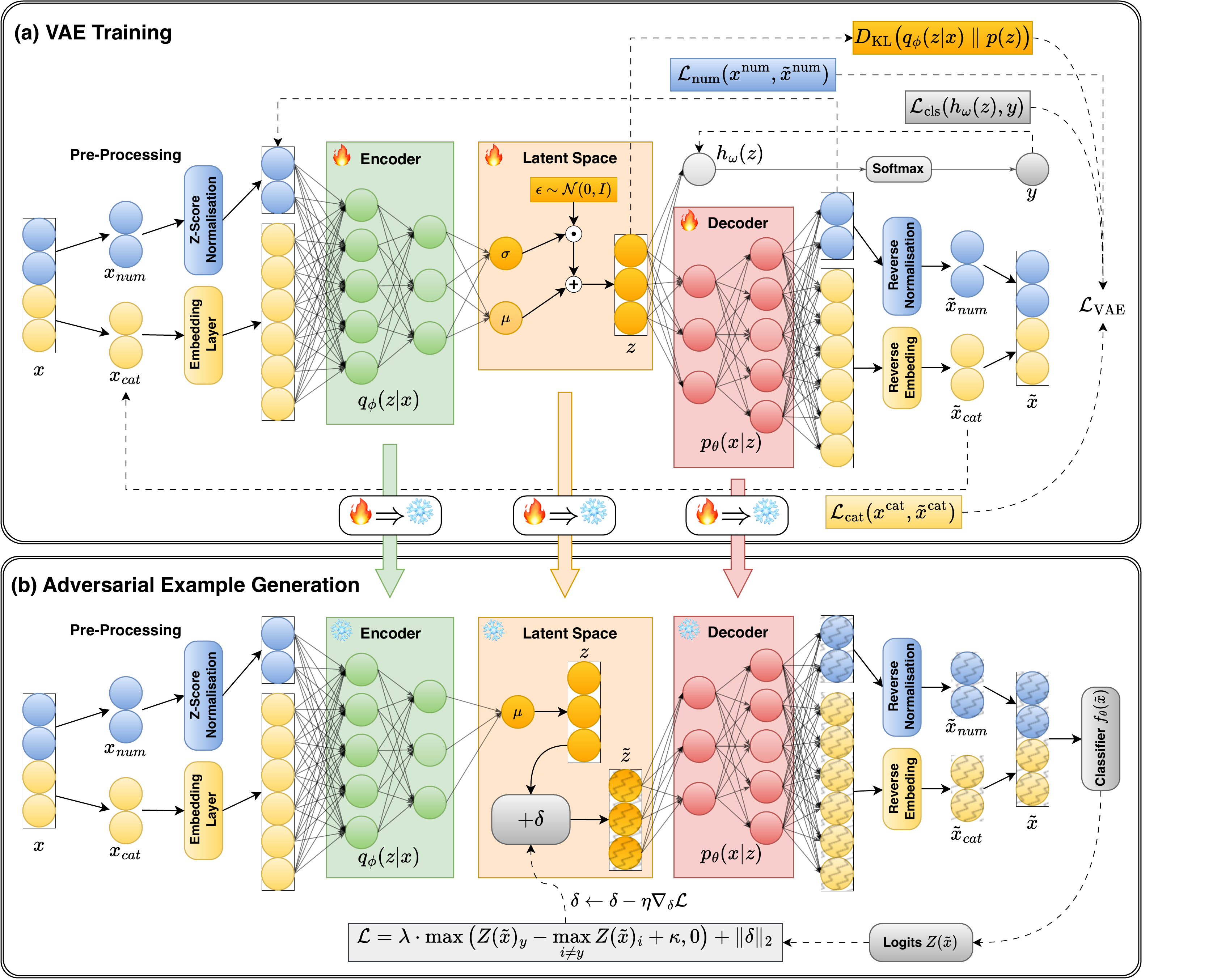}
    \caption{Overall framework of the proposed VAE-based adversarial example generation. \textbf{(a) VAE Training:} Input features $x$ (categorical and numerical) are pre-processed via embedding and normalisation before being encoded into latent variables $(\mu, \sigma)$. The decoder reconstructs both numerical and categorical features under the joint VAE loss. An auxiliary classification head $h_\omega(z)$, shown as a grey node, is employed during training to promote a discriminative and well-structured latent space; it is not used during adversarial generation. \textbf{(b) Adversarial Example Generation:} After training, the encoder and decoder are frozen. For each input, a latent representation $z$ is obtained and iteratively perturbed by an instance-specific vector $\delta$ optimised via gradient descent on the C\&W-style attack objective. The decoder reconstructs the perturbed latent vector into an adversarial example $\tilde{x}=p_\psi(z+\delta^*)$, which remains on the data manifold while manipulating the classifier's logits $Z(\tilde{x})$ to cross the decision boundary.}
    \label{fig:overall-design}
\end{figure*}

The objective is to generate adversarial examples $\tilde{x}$ that satisfy two criteria: (1) proximity to the original input $x$ under a distance metric $d(\tilde{x}, x)$, and (2) misclassification by the classifier, i.e., $f_\theta(\tilde{x}) \neq f_\theta(x)$. To address the discrete nature of categorical features, we leverage a VAE in Figure~\ref{fig:overall-design} to learn a latent representation $z_i \in \mathbb{R}^k$ that unifies both $x_i^{\text{cat}}$ and $x_i^{\text{num}}$ into a continuous space. VAEs are theoretically well-suited for this task due to their variational inference framework, which can ensure the latent space is both compact and semantically meaningful~\citep{kingma2013auto,apellaniz2024improved,wang2025ttvae}. This is critical to generate perturbations that preserve feature relationships. Adversarial perturbations are then applied in this latent space to craft adversarial examples for tabular data. In contrast to other generative models like GANs~\citep{Anshelevich2025Synthetic,fonseca2023tabular}, which lack explicit density modelling, VAEs provide a principled trade-off between reconstruction fidelity and latent space structure, enabling controllable perturbation strategies aligned with the data distribution.




\subsection{VAE with Classification Loss}
The VAE is designed to learn a meaningful latent representation of the tabular input data. The VAE architecture comprises an encoder, a decoder, and a classification head, as shown in Figure~\ref{fig:overall-design}(a).

\paragraph{Encoder $q_\phi(z|x)$} The encoder maps $x = [x^{\text{cat}}, x^{\text{num}}]$ to a latent distribution $z \sim q_\phi(z|x)$. Categorical features $x^{\text{cat}}$ are first embedded into dense vectors using an embedding layer, following established practices for mixed-type tabular data~\citep{zhang2023mixed}, while numerical features $x^{\text{num}}$ are processed through fully connected layers. These transformed representations are concatenated and passed through shared encoder layers consisting of multiple fully connected layers with batch normalisation and ReLU activations. The final hidden representation is then projected onto the parameters of a Gaussian latent distribution, characterised by mean $\mu_z$ and variance $\sigma_z^2$. The latent vector $z$ is then obtained using the reparameterisation trick~\citep{kingma2013auto}, $z = \mu_z + \sigma_z \odot \epsilon$, where $\epsilon \sim \mathcal{N}(0, I)$.

\paragraph{Decoder $p_\psi(x|z)$} The decoder reconstructs the input from the latent representation $z$, generating $\tilde{x} = [\tilde{x}^{\text{cat}}, \tilde{x}^{\text{num}}]$. For categorical reconstruction, $\tilde{x}^{\text{cat}}$ outputs softmax probabilities over the categories of each feature. During inference, these probabilities are converted to discrete labels via \( \arg\max \) to recover the original label-encoded format. For numerical features, $\tilde{x}^{\text{num}}$ regresses numerical values using fully connected layers under a Gaussian likelihood assumption.

\paragraph{Classifier $h_\omega(z)$} The classification head, operating directly on the latent space, predicts labels $y$ through loss $\mathcal{L}_{\text{cls}}(h_\omega(z), y)$. Since our adversarial attacks are untargeted, we do not specify a target class, so label conditioning is unnecessary. Hence, we forgo a conditional VAE~\citep{joy2020capturing} that enforces a particular label. Instead, a simpler VAE with a classification head suffices to separate different classes of input in latent space.

\paragraph{Loss Function $\mathcal{L}_{\text{VAE}}$} The VAE is trained by minimising a composite loss function:

\begin{align}\label{eq:vae-loss}
    \mathcal{L}_{\text{VAE}} &= 
    \underbrace{
        \mathbb{E}_{q_\phi(z|x)} \Big[ 
            \underbrace{\|x^{\text{num}} - \tilde{x}^{\text{num}}\|^2}_{\substack{\text{Numerical reconstruction} \\ \text{(MSE, Gaussian likelihood)}}} 
            + 
            \underbrace{\mathcal{L}_{\text{cat}}(x^{\text{cat}}, \tilde{x}^{\text{cat}})}_{\substack{\text{Categorical reconstruction} \\ \text{(Cross-entropy, Softmax probability)}}} 
        \Big]
    }_{\text{Reconstruction Loss}} \nonumber \\
    &\quad + \beta \cdot \underbrace{D_{\mathrm{KL}}\big(q_\phi(z|x) \parallel p(z)\big)}_{\substack{\text{KL divergence} \\ \text{(Latent regularisation)}}} 
    + \alpha \cdot \underbrace{\mathcal{L}_{\text{cls}}(h_\omega(z), y)}_{\substack{\text{Classification loss} \\ \text{(Class separation)}}},
\end{align}

where $\mathcal{L}_{\text{cat}}$ computes the categorical cross-entropy loss across all $m$ categorical features and $\mathcal{L}_{\text{cls}}$ is the classification cross-entropy loss. The KL divergence term $D_{\mathrm{KL}}$ enforces alignment between the encoder's posterior $q_\phi(z|x)$ and a standard Gaussian prior $p(z) = \mathcal{N}(0, I)$. Hyperparameters $\alpha$ and $\beta$ modulate the contributions of the classification loss and KL regularisation, respectively.

\paragraph{Training Process}
The training procedure jointly optimises the encoder parameters $\phi$, decoder parameters $\psi$, and classifier parameters $\omega$. The reconstruction loss ensures accuracy of the decoded numerical and categorical features to the original input, employing mean squared error (MSE) for numerical features and cross-entropy for categorical features. The KL-divergence term regularises the latent space by encouraging proximity to the Gaussian prior, creating a smooth, continuous latent manifold where small perturbations correspond to semantically meaningful variations rather than arbitrary noise. This regularisation is implemented via the reparameterisation trick to enable gradient-based optimisation, while ensuring that adversarial perturbations in latent space translate to realistic data variations. Simultaneously, the classification loss $\mathcal{L}_{\text{cls}}$ incentivises the latent representation $z$ to encode discriminative information for the predictive task, ensuring that adversarial examples can effectively exploit decision boundaries. This unified training strategy yields a structured and task-aware latent space, facilitating effective adversarial example generation through gradient-guided perturbations that leverage the manifold structure to maintain data realism while maximising attack success.


\paragraph{Gradient Flow During Adversarial Optimisation}
During adversarial example generation (Section~\ref{subsec:adv-gen}), the encoder parameters $\phi$ are frozen. Gradients from the classification loss $\mathcal{L}_{\text{cls}}(h_\omega(z), y)$ are backpropagated to the latent variable $z$ only, not to $\phi$. This design ensures that adversarial perturbations are applied within the learned latent manifold without altering the encoder's mapping, thereby preserving reconstruction fidelity and latent-space consistency across attacks.

\subsection{Adversarial Example Generation}\label{subsec:adv-gen}


\paragraph{Perturbation Calculation}
Adversarial examples are generated by introducing an additive noise vector $\delta \in \mathbb{R}^k$ to the latent vector $z$ of an input $x$ in the latent space, as shown in Figure~\ref{fig:overall-design}(b). The perturbation $\delta$ is optimised to induce misclassification while remaining constrained to a small neighbourhood around the original latent code, ensuring the adversarial example $\tilde{x}$ stays on the data manifold. According to the Carlini \& Wagner (C\&W) paradigm~\citep{carlini2017towards}, the attack is achieved by solving the following optimisation problem:

\begin{equation}\label{eq:cw-obj}
    \delta^* = \arg\min_{\delta} \bigg[ \lambda \cdot \max\Big( Z(\tilde{x})_y - \max_{i \neq y} Z(\tilde{x})_i + \kappa, 0\Big) + \|\delta\|_2 \bigg],
\end{equation}

where $Z(\tilde{x})$ denotes the logits (pre-softmax outputs) of the classifier $f_\theta$
for the reconstructed adversarial example $\tilde{x} = p_\psi(z + \delta^*)$, $y$ is the original true label, and $\kappa \geq 0$ controls the desired confidence margin for misclassification. The $\ell_2$ norm term $\|\delta\|_2$ constrains latent space perturbations while the scalar $\lambda$ balances the trade-off between the attack's success and the size of the perturbation. 
This optimisation is performed individually for each input instance through iterative gradient-based updates on the latent perturbation $\delta$, rather than by a separately trained generator or module.

\begin{algorithm}[t]
    \caption{VAE-Based Adversarial Example Generation}
    \label{alg:vae_adversarial}
    \textbf{Input}: Input data $\mathcal{D} = \{(x_i, y_i)\}_{i=1}^N$, classifier $f_\theta$ with logits $Z(\cdot)$, 
    VAE $(q_\phi, p_\psi, h_\omega)$ \\
    \textbf{Parameters}: Attack hyperparameters $\lambda, \kappa$, iterations $T$, learning rate $\eta$, convergence threshold $\tau$ \\
    \textbf{Output}: Adversarial examples $\{\tilde{x}_i\}_{i=1}^N$ 

    \begin{algorithmic}[1]
        \STATE Train VAE: Minimise $\mathcal{L}_{\text{VAE}}$ (Eq.~\ref{eq:vae-loss}) to learn encoder $q_\phi$, decoder $p_\psi$, and classifier $h_\omega$
        \FOR{each input $(x, y) \in \mathcal{D}$}
            \STATE Encode $x$ to latent mean: $z \leftarrow \mu_z(x) = \mathbb{E}[q_\phi(z|x)]$
            \STATE Initialise perturbation: $\delta \leftarrow \mathbf{0}$ 
            \FOR{$t = 1$ \TO $T$}
                \STATE Decode perturbed latent: $\tilde{x} \leftarrow p_\psi(z + \delta)$
                \STATE Compute logits: $Z(\tilde{x}) \leftarrow$ classifier logits of $f_\theta(\tilde{x})$
                \STATE Compute loss:
                \[
                \mathcal{L} = \lambda \cdot \max\big( Z(\tilde{x})_y - \max_{i \neq y} Z(\tilde{x})_i + \kappa, 0 \big) 
                + \|\delta\|_2
                \]
                \STATE Gradient update: $\delta \leftarrow \delta - \eta \nabla_\delta \mathcal{L}$
                \IF{$\|\delta^{(t)} - \delta^{(t-1)}\|_2 < \tau$} 
                \STATE \textbf{break} 
                \ENDIF 
            \ENDFOR
            \STATE Generate adversarial example: $\tilde{x} \leftarrow p_\psi(z + \delta)$
        \ENDFOR
        \STATE \textbf{return} $\{\tilde{x}\}$
    \end{algorithmic}
\end{algorithm}

\paragraph{VAE-based Attack Algorithm}
As depicted in Algorithm~\ref{alg:vae_adversarial}, the optimisation proceeds through three phases. First, the input $x$ is encoded to its latent representation $z = \mu_z(x)$ using the mean of the encoder distribution $q_\phi(z|x)$. Next, gradient-based optimisation (e.g., Adam) iteratively adjusts $\delta$ to minimise Equation~\ref{eq:cw-obj}, where the logit margin loss drives the classifier's prediction away from the true class $y$. Finally, the optimised perturbation $\delta^*$ is decoded through $p_\psi$ to produce the adversarial example $\tilde{x} = p_\psi(z + \delta^*)$. This approach ensures adversarial examples remain on the data manifold through the VAE's generative constraints while directly targeting the classifier's decision boundaries through logit manipulation.

\paragraph{Deterministic vs. Stochastic Latent Encodings}
In Algorithm~\ref{alg:vae_adversarial}, we use the latent mean $z = \mu_z(x)$ rather than sampling $z \sim q_\phi(z|x)$ to initialise adversarial optimisation. This deterministic choice ensures stable and reproducible attack trajectories by avoiding variance introduced by random sampling. Since the encoder distribution $q_\phi(z|x)$ is typically narrow after VAE training, $\mu_z(x)$ provides a representative latent code lying near high-density regions of the manifold. We therefore use the deterministic mean for stable and reproducible optimisation; we leave a stochastic Monte-Carlo variant (averaging gradients over multiple $z\sim q_\phi(z|x)$) as future work.

\section{Evaluation}\label{sec:eval}

To evaluate the performance and robustness of our proposed method, we designed experiments to address three research questions (RQs):

\begin{itemize}
    \item \textbf{RQ1: How effectively does the VAE reconstruct input data across categorical and numerical features?}
    \item \textbf{RQ2: How effective and imperceptible are VAE-based adversarial examples compared to traditional input-space attacks and other VAE-based methods?}
    \item \textbf{RQ3: How do hyperparameters, sparsity constraints, and generative model choices affect VAE-based adversarial attack performance?}
\end{itemize}

Each task in our experimental design corresponds to one of these research questions:

\paragraph{Task 1: Reconstruction and Latent Representation Evaluation (addressing RQ1)}  
We assess the VAE's capability to reconstruct mixed-type tabular data while preserving latent discriminability through three interconnected evaluations. 

First, predictive performance is rigorously tested by training downstream classifiers on reconstructed inputs \( \tilde{x} \) and comparing their accuracy to models trained on original data~\( x \). The predictive accuracy delta \( \delta_{\text{acc}} = \text{Acc}(x) - \text{Acc}(\tilde{x}) \) serves as a critical indicator of latent space information retention, with smaller values indicating better preservation of class-discriminative patterns. Additionally, we report the accuracy retention ratio \( A_{\text{ret}} = 1 - \text{Acc}(\tilde{x})/\text{Acc}(x) \), which expresses the proportion of baseline accuracy retained after reconstruction, with higher values corresponding to smaller accuracy losses and stronger latent-space information preservation. The two complementary metrics jointly describe reconstruction quality from both absolute and relative perspectives.

Second, categorical reconstruction quality is measured via the average accuracy \( x^{\text{cat}}_{\text{acc}} \), which quantifies the proportion of correctly reconstructed label-encoded categories across all \( m \) categorical features. For numerical features, we employ a triad of complementary metrics: the coefficient of determination \( R^2 \) evaluates variance preservation, cosine similarity \( \cos(x^{\text{num}}, \tilde{x}^{\text{num}}) \) verifies directional alignment, and Pearson correlation \( \rho(x^{\text{num}}, \tilde{x}^{\text{num}}) \) confirms linear distributional consistency between original and reconstructed features.  

Finally, latent representation quality is analysed through t-SNE visualisation of class separation and controlled ablation studies, where removing the classification loss \( \mathcal{L}_{\text{cls}} \) (\( \alpha = 0 \)) isolates its impact on reconstruction quality and class separability.

\paragraph{Task 2: Adversarial Attack Effectiveness and Imperceptibility (addressing RQ2)}  
We evaluate adversarial attacks through two key criteria: (1) \textbf{effectiveness} in inducing misclassifications, measured by Attack Success Rate (ASR), and (2) \textbf{imperceptibility}, quantified via statistical consistency with the data manifold. 

ASR is defined as: $\frac{1}{N}\sum_{i=1}^N \mathbb{I}(f_\theta(\tilde{x}_i) \neq y_i)$, where $\mathbb{I}$ is the indicator function. 

To measure imperceptibility in the latent space, deviations~\citep{he2025investigating} are quantified via the Mahalanobis distance (MD):
\[
\text{MD}(z + \delta) = \sqrt{(z + \delta - \mu_z)^\top \Sigma_z^{-1} (z + \delta - \mu_z)},
\] 
where \(z = \mu_z(x)\) is the latent code of the original input \(x\), and \(\mu_z\), \(\Sigma_z\) are the mean and covariance matrix of the training data's latent distribution \(q_\phi(z|x)\). Larger MD values indicate greater deviation from the manifold of training samples. 

To statistically identify outliers, we apply a $\chi^2$ test at the 95\% confidence level, where adversarial examples are flagged as outliers if \(\text{MD}^2(z + \delta) > \chi^2_{0.95}(k)\), with \(k\) being the number of dimensions in the latent space. Then, we compute the \textbf{Outlier Rate (\(\mathcal{O}_r\))} as:
\[
\mathcal{O}_r = \frac{1}{N}\sum_{i=1}^N \mathbb{I}\left(\text{MD}^2(z_i + \delta_i) > \chi^2_{0.95}(k)\right).
\]

We introduce the \textbf{In-Distribution Success Rate (IDSR)} as a metric to comprehensively evaluate the percentage of successful adversarial attacks that also remain within the original data distribution. It is defined as:
$\text{IDSR} = \text{ASR} \times (1 - \mathcal{O}_r)$.


\paragraph{Task 3: Performance Factors and Method Analysis (addressing RQ3)}  

We investigate multiple factors that influence VAE-based adversarial attack performance, including hyperparameter sensitivity, sparsity control mechanisms, architectural choices, and method limitations.

\begin{itemize}
    \item \textbf{Hyperparameter Analysis:} We evaluate the sensitivity of attack performance to critical hyperparameters: the regularization weight \(\lambda\), which balances attack success rate (ASR) and perturbation magnitude, and the learning rate \(\eta\), which governs optimization stability. Experiments span \(\lambda \in [0.1, 1]\) and \(\eta \in \{0.01, 0.03, 0.1, 0.15, 0.2, 0.25, 0.3\}\), evaluating their impact on ASR and mean \(\ell_2\) perturbation distance in the latent space. 
    \item \textbf{Sparsity Control Analysis:} We explore explicit sparsity constraints to minimise the number of modified features while maintaining attack effectiveness. We investigate three approaches: (1) differentiable $\ell_0$ approximations using sigmoid-based smoothing functions, (2) greedy search-based feature selection for post-hoc sparsity optimisation, and (3) $\ell_1$ norm regularisation as a convex relaxation of the $\ell_0$ constraint~\citep{tibshirani1996regression}. Performance is evaluated using sparsity rate (percentage of modified features) and $\ell_0$ norm (number of changed features).
    \item \textbf{Generative Architectural Comparison:} We compare VAE and GAN architectures for tabular data reconstruction to validate our generative model choice. Reconstruction quality is assessed using the same metrics as Task 1, focusing on how architectural differences impact adversarial generation capability.
\end{itemize}

\subsection{Datasets}

We evaluate our method on six tabular datasets spanning different domains and feature types. Table~\ref{tab:datasets} summarizes their key characteristics. All datasets undergo standardised preprocessing to ensure compatibility across models. Numerical features are normalised using Z-score standardisation, while binary features are encoded as $\{0, 1\}$. Categorical features employ integer label encoding to feed the embedding layers. The data is partitioned into training (70\%), validation (10\%), and test (20\%) sets using stratified splitting to maintain class distribution.

\begin{table}[t]
\caption{Dataset characteristics: $N$ = instances, $d$ = total features, $d_{\text{num}}$ = numerical features, $d_{\text{cat}}$ = categorical features, $d_{\text{binary}}$ = binary features (subset of categorical), and $y$ = number of classes.}
\label{tab:datasets}
\centering
\begin{tabular}{@{}lrrrrrr@{}}
\toprule
\textbf{Dataset} & \textbf{$N$} & \textbf{$d$} & \textbf{$d_{\text{num}}$} & \textbf{$d_{\text{cat}}$} & \textbf{$d_{\text{binary}}$} & \textbf{$y$} \\ \midrule
Adult       & 30,162  & 12 & 4  & 8  & 1  & 2 \\
Phishing    & 11,430  & 86 & 57 & 29 & 28 & 2 \\   
Pendigits   & 10,992  & 16 & 16 & 0  & 0  & 10 \\
German      & 1,000   & 19 & 8  & 11 & 2  & 2  \\
Electricity & 45,312  & 8  & 7  & 1  & 0  & 2  \\
Covertype   & 58,1012 & 54 & 10 & 54 & 54 & 7  \\
\bottomrule
\end{tabular}
\end{table}

The datasets provide diverse evaluation scenarios for tabular adversarial attacks. For the \textbf{Adult} dataset, the model predicts income levels using census data with mixed categorical and numerical features, representing classic tabular prediction tasks. For the \textbf{Phishing} dataset, the model detects malicious websites through URL and content features~\cite{hannousse2021towards}, providing a high-dimensional dataset with predominantly binary categorical features. For the \textbf{Pendigits} dataset, the model recognises handwritten digits from pen trajectory coordinates, offering a purely numerical multiclass scenario. For the \textbf{German} credit dataset, the model predicts loan default risk with balanced categorical-numerical composition but limited sample size. For the \textbf{Electricity} dataset, the model forecasts electricity price changes using temporal market features with minimal categorical content. Finally, for the \textbf{Covertype} dataset, the model predicts forest cover types from cartographic features, combining moderate numerical complexity with extensive categorical variables in a multiclass setting.





\subsection{Predictive Models}

We evaluate our proposed VAE-based adversarial attack method using three distinct predictive models widely adopted for tabular data: a Multi-Layer Perceptron (MLP), a Soft Decision Tree (SDT)~\citep{frosst2017distilling}, and a TabTransformer~\citep{huang2020tabtransformer}. These models were chosen to represent a range of architectures, from traditional neural networks to advanced transformer-based approaches.

\paragraph{Multi-Layer Perceptron (MLP)}
The MLP serves as a baseline neural architecture for tabular data, comprising fully connected layers with nonlinear activation functions (e.g., ReLU). It learns hierarchical feature representations through standard backpropagation, optimising task-specific objectives such as cross-entropy loss for classification. 

\paragraph{Soft Decision Tree (SDT)}
The SDT generalises classical decision trees by replacing deterministic splits with differentiable, probabilistic decisions. Each internal node applies a sigmoid-activated linear layer to route samples through branches, enabling end-to-end gradient-based optimisation. 

\paragraph{TabTransformer}
The TabTransformer adapts the transformer architecture for tabular data by embedding individual features and processing them through self-attention mechanisms. Unlike vision or language transformers, it employs feature-wise embeddings to preserve column-specific semantics and uses attention to model interactions between features.

\subsection{VAE Configuration}

\begin{table}[!htbp]
\caption{VAE hyperparameters that vary across datasets. Common settings: learning rate = 1e-2, batch size = 512.}
\label{tab:vae_hyperparams}
\centering
\begin{tabular}{@{}lrrrr@{}}
\toprule
\textbf{Dataset} & \textbf{Encode/Decode} & \textbf{Latent Dim} & \textbf{Epochs} & \textbf{$\beta$} \\ \midrule
Adult       & [128, 64] & 16 & 200 & 1e-3 \\
Phishing    & [128, 64] & 16 & 200 & 1e-2 \\
Pendigits   & [64, 32]  & 8  & 200 & 1e-2 \\
German      & [64, 32]  & 8  & 100 & 1e-2 \\
Electricity & [16]      & 8  & 100 & 1e-4 \\
Covertype   & [128, 64] & 16 & 50  & 1e-4 \\
\bottomrule
\end{tabular}
\end{table}

Table~\ref{tab:vae_hyperparams} presents the dataset-specific hyperparameters selected through grid search to maximise reconstruction performance on validation sets. The overall VAE architecture is kept consistent across datasets, with only key hyperparameters (latent dimension, layer width, and $\beta$) tuned to maintain comparable reconstruction quality rather than to explore architectural variations. All models use the Adam optimiser with a learning rate of 1e-2 and batch size of 512. The encoder and decoder architectures are symmetric, using ReLU activation functions between layers. The loss function combines reconstruction losses for categorical and numerical features with a Kullback-Leibler (KL) divergence term weighted by $\beta$. We set $\alpha=1$ to enable the classification loss ($\mathcal{L}_{\text{cls}}$) as defined in Eq.~(\ref{eq:vae-loss}). 

The VAE is trained on the training and validation sets of each dataset to ensure it captures the underlying distribution of the data. The trained VAE is then used to generate adversarial examples by manipulating the learned latent representations.

\subsection{Attack Methods}
 
\paragraph{Input-space Baseline} We compare against two standard adversarial attacks adapted for tabular data as reference baselines:

\begin{itemize}
    \item \textbf{FGSM}~\citep{goodfellow2015explaining}: A single-step attack generating adversarial examples as
\(\tilde{x} = x + \epsilon \cdot \text{sign}(\nabla_x \mathcal{L}(f_\theta(x), y))\), where $\epsilon$ bounds the $L_\infty$ perturbation magnitude. We set $\epsilon = 0.5$ (50\% of feature ranges) through grid search.
    \item \textbf{PGD}~\citep{madry2017towards}: An iterative $L_\infty$ attack with \(\tilde{x}^{(t+1)} = \text{Clip}_\epsilon\left(\tilde{x}^{(t)} + \alpha \cdot \text{sign}(\nabla_x \mathcal{L}(f_\theta(\tilde{x}^{(t)}), y))\right)\), using $T=10$ iterations and step size $\alpha = \epsilon/T$ following standard practice.  
\end{itemize}

Both baselines operate directly in \textit{input space}, unlike our latent-space approach. While effective in continuous domains like images, they face challenges with tabular data: discrete features (binary/categorical) require post-hoc rounding, and perturbations often violate feature correlations or domain constraints (e.g., unrealistic age values). 

\paragraph{VAE-based Baseline Methods}
Since there are no existing VAE-based adversarial attack methods specifically designed for tabular data, we adapt two VAE attack approaches from the image domain to serve as baselines. These methods are implemented by replacing the attack mechanism while using the same VAE architecture and training described in the previous section. 

\begin{itemize}
    \item \textbf{PGD-VAE}~\citep{stutz2019disentangling} applies the PGD algorithm directly in the VAE's latent space using the same $\epsilon$ and iteration constraints as traditional PGD, adapted for latent space optimisation. 
    \item \textbf{DeltaZ}~\citep{creswell2017latentpoison} implements a multiplicative perturbation method that transforms latent vectors as $z \cdot (1 + \Delta z)$, where $\Delta z$ is optimized to flip feature signs in dimensions that encode class-discriminative information. DeltaZ is specifically designed for binary classification tasks and assumes that opposite classes have features with different signs in the latent space. 
\end{itemize}

\paragraph{Our VAE-based Attack}
The adversarial example generation process follows Algorithm~\ref{alg:vae_adversarial}, with perturbations $\delta$ optimised in the VAE's latent space using gradient descent. Key hyperparameters include a maximum iteration limit $T = 300$, learning rate $\eta = 0.1$, regularisation coefficient $\lambda = 1$, convergence threshold $\tau = 10^{-5}$, and confidence margin $\kappa = 0$. The loss function balances two objectives: a \textit{margin-based misclassification term} that reduces the confidence of the original class below other classes (untargeted attack) and an \textit{$L_2$-regularisation term} that penalises large latent-space deviations to preserve imperceptibility in proximity. Perturbations are initialised at $\delta = \mathbf{0}$ and iteratively refined until either the change in $\delta$ falls below $\tau$ (indicating convergence) or the iteration limit $T$ is reached, whichever occurs first. This configuration adapts the Carlini \& Wagner attack~\citep{carlini2017towards} to latent-space optimisation, enforcing manifold-aligned perturbations through the VAE's decoder. We set the confidence margin $\kappa=0$ to prioritise minimal, manifold-preserving perturbations\footnote{The confidence margin $\kappa$ in the attack loss controls how strongly the adversarial example is required to exceed the decision boundary: larger $\kappa$ increases attack confidence but typically demands larger perturbations.}, ensuring the optimiser seeks the smallest latent displacement sufficient to flip the prediction.
All parameters remain fixed across datasets and models to ensure consistent evaluation of effectiveness and imperceptibility of attacks.


\section{Results and Discussion}
In this section, the results are structured around the task defined in Section \ref{sec:method}, followed by an in-depth discussion.

\subsection{Results for Task 1: Reconstruction and Latent Representation Evaluation (RQ1)}

\begin{table}[t]
\centering
\caption{Reconstruction performance comparison between the proposed VAE (with $\mathcal{L}_{\text{cls}}$) and baseline VAE (without $\mathcal{L}_{\text{cls}}$). Metrics include accuracy delta ($\delta_{\text{acc}}$), accuracy retention ratio ($A_\text{ret}$), MSE, coefficient of determination ($R^2$), cosine similarity ($\cos$), Pearson correlation ($\rho$), and categorical reconstruction accuracy ($x^{\text{cat}}_{\text{acc}}$). Best results are in \textbf{bold}.}
\label{tab:vae-recon-results}
\begin{tabular}{@{}lcrrrrrrr@{}}
\toprule
Dataset & VAE & $\delta_{acc}$ & $A_\text{ret}$ & MSE & $R^2$ & $cos$ & $\rho$ & $x^{cat}_{acc}$ \\ \midrule
Adult & c $\mathcal{L}_{\text{cls}}$ & \textbf{0.0030} & \textbf{0.9965} & \textbf{0.0308} & \textbf{0.9679} & \textbf{0.9851} & \textbf{0.9854} & 0.9864 \\
 & w/o $\mathcal{L}_{\text{cls}}$ & 0.0056 & 0.9934 & 0.0440 & 0.9542 & 0.9776 & 0.9777 & \textbf{0.9870} \\
  \addlinespace
Phishing & c $\mathcal{L}_{\text{cls}}$ & \textbf{0.0184} & \textbf{0.9809} & \textbf{0.1487} & \textbf{0.8320} & \textbf{0.9124} & \textbf{0.9130} & \textbf{0.9552} \\
 & w/o $\mathcal{L}_{\text{cls}}$ & 0.0324 & 0.9663 & 0.2108 & 0.7618 & 0.8742 & 0.8743 & 0.9529 \\
 \addlinespace
German & c $\mathcal{L}_{\text{cls}}$ & \textbf{0.0100} & \textbf{0.9866} & \textbf{0.2164} & \textbf{0.7878} & \textbf{0.8888} & \textbf{0.8892} & \textbf{0.7388} \\
 & w/o $\mathcal{L}_{\text{cls}}$ & 0.0350 & 0.9530 & 0.2517 & 0.7533 & 0.8691 & 0.8694 & 0.7215 \\
 \addlinespace
Electricity & c $\mathcal{L}_{\text{cls}}$ & \textbf{0.0276} & \textbf{0.9660} & \textbf{0.0658} & \textbf{0.9438} & \textbf{0.9719} & \textbf{0.9721} & 0.9993 \\
 & w/o $\mathcal{L}_{\text{cls}}$ & 0.0389 & 0.9520 & 0.0734 & 0.9373 & 0.9687 & 0.9705 & \textbf{0.9996} \\
  \addlinespace
Covertype & c $\mathcal{L}_{\text{cls}}$ & \textbf{0.0419} & \textbf{0.9499} & \textbf{0.0296} & \textbf{0.9705} & \textbf{0.9862} & \textbf{0.9872} & \textbf{0.9999} \\
 & w/o $\mathcal{L}_{\text{cls}}$ & 0.0512 & 0.9387 & 0.0496 & 0.9504 & 0.9759 & 0.9762 & \textbf{0.9999} \\
 \addlinespace
PenDigits & c $\mathcal{L}_{\text{cls}}$ & \textbf{0.0059} & \textbf{1.0023} & \textbf{0.0588} & \textbf{0.9417} & \textbf{0.9706} & \textbf{0.9707} & N/A \\
 & w/o $\mathcal{L}_{\text{cls}}$ & 0.0254 & 0.9741 & 0.0641 & 0.9364 & 0.9677 & 0.9678 & N/A \\
\bottomrule
\end{tabular}%
\end{table}

Table~\ref{tab:vae-recon-results} demonstrates the consistent superiority of our VAE with classification loss ($\mathcal{L}_{\text{cls}}$) as evidenced by (1) superior preservation of downstream predictive performance, (2) enhanced numerical feature reconstruction, and (3) robust categorical reconstruction. An ablation study confirms these benefits by demonstrating that removing $\mathcal{L}_{\text{cls}}$ (setting $\alpha = 0$) degrades both reconstruction quality and latent discriminability across all evaluated datasets.

\paragraph{Predictive Ability Preservation}  

The proposed model consistently reduces accuracy deltas ($\delta_{\text{acc}}$) by 43--82\% compared to the baseline across all six datasets, demonstrating superior preservation of class-discriminative information. 
In addition, we report the accuracy retention ratio $A_\text{ret}$ as a normalised complement to $\delta_{\text{acc}}$, providing a scale-independent measure of predictive ability. Across all datasets, $A_\text{ret}$ remains above 0.95 for the proposed model, and in several cases (e.g., \emph{PenDigits}) slightly exceeds 1.0, indicating reconstructed inputs occasionally yield comparable or marginally higher predictive scores than originals due to regularisation effects.
The improvements are most pronounced on datasets with complex feature relationships, where classification loss provides stronger guidance for maintaining discriminative patterns in the latent space. Even the German dataset, which shows the highest absolute $\delta_{\text{acc}}$ values due to reconstruction difficulties, benefits significantly from the classification loss. These improvements confirm that the classification loss $\mathcal{L}_{\text{cls}}$ enforces latent representations that maintain downstream predictive performance, with classifiers trained on reconstructed data achieving near-original performance levels.

\paragraph{Numerical Feature Reconstruction Quality}  

Our approach achieves consistent improvements across all numerical metrics: 8--40\% lower $MSE$, 1--10\% higher $R^2$ values, and consistently improved cosine similarity. The improvements are consistent across datasets with varying complexity and feature distributions, with Pearson correlations ranging from 0.889 to 0.987 for all datasets with $\mathcal{L}_{\text{cls}}$. This confirms that joint optimisation of reconstruction and classification objectives enhances numerical feature alignment while preserving linear relationships in the latent space.

\paragraph{Categorical Feature Reconstruction Quality}  

The models achieve exceptional categorical reconstruction accuracy over most datasets exceeding 95\% regardless of $\mathcal{L}_{\text{cls}}$, with minimal differences between approaches. However, over the German dataset the VAE presents a notable exception with substantially lower categorical accuracy ($\sim 74\%$) due to its small dataset size ($N=1000$) in Table~\ref{tab:datasets}, which is insufficient for effective VAE training~\citep{kiran2023comparative}. This limited training data results in poor reconstruction quality that affects overall VAE performance.


\begin{figure}[t]
    \centering
    \begin{subfigure}[t]{0.48\columnwidth}
        \centering
        \includegraphics[width=\textwidth]{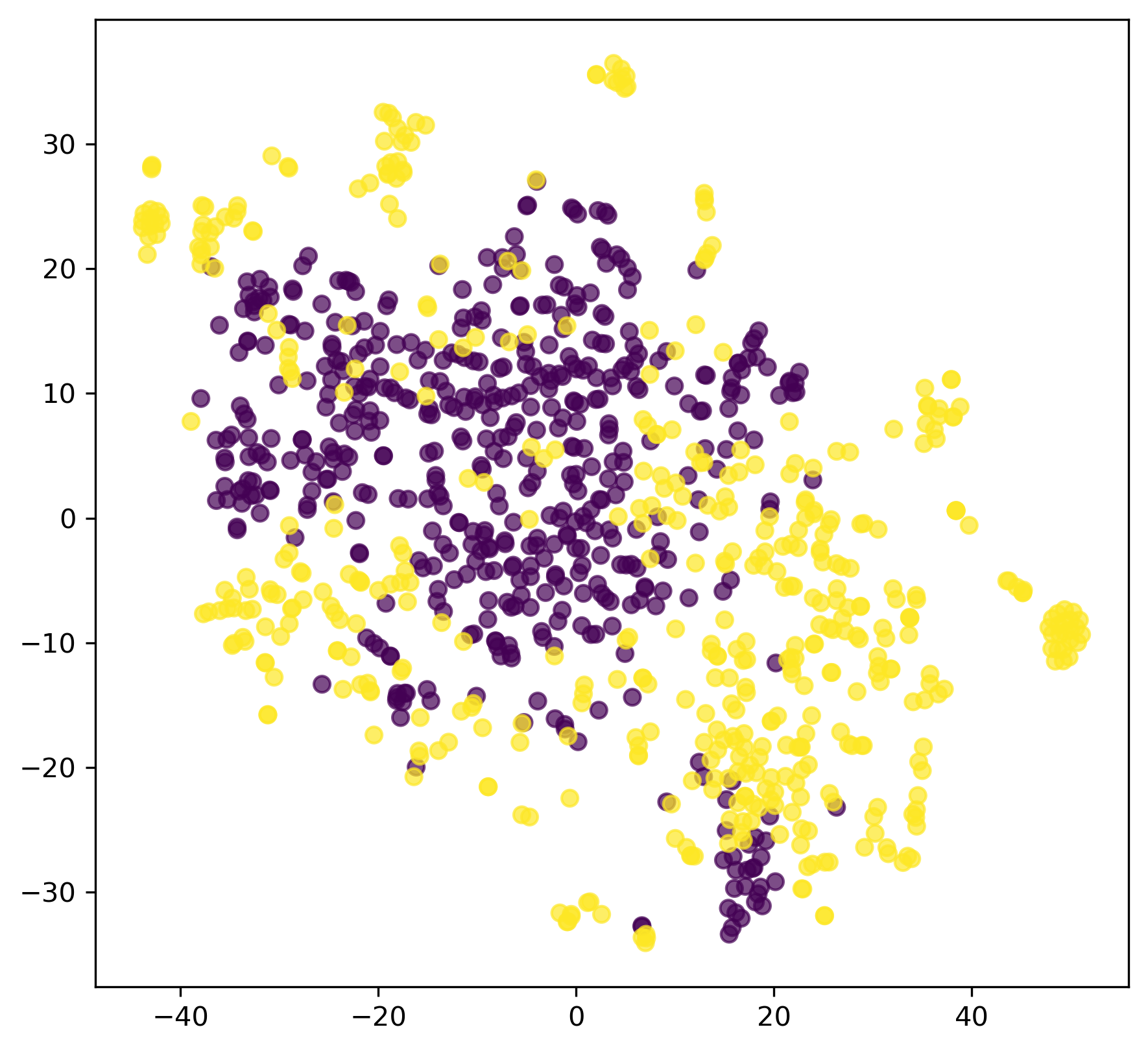}
        \caption{VAE without $\mathcal{L}_{\text{cls}}$}
        \label{fig:tsne_0}
    \end{subfigure}
    \hfill
    \begin{subfigure}[t]{0.48\columnwidth}
        \centering
        \includegraphics[width=\textwidth]{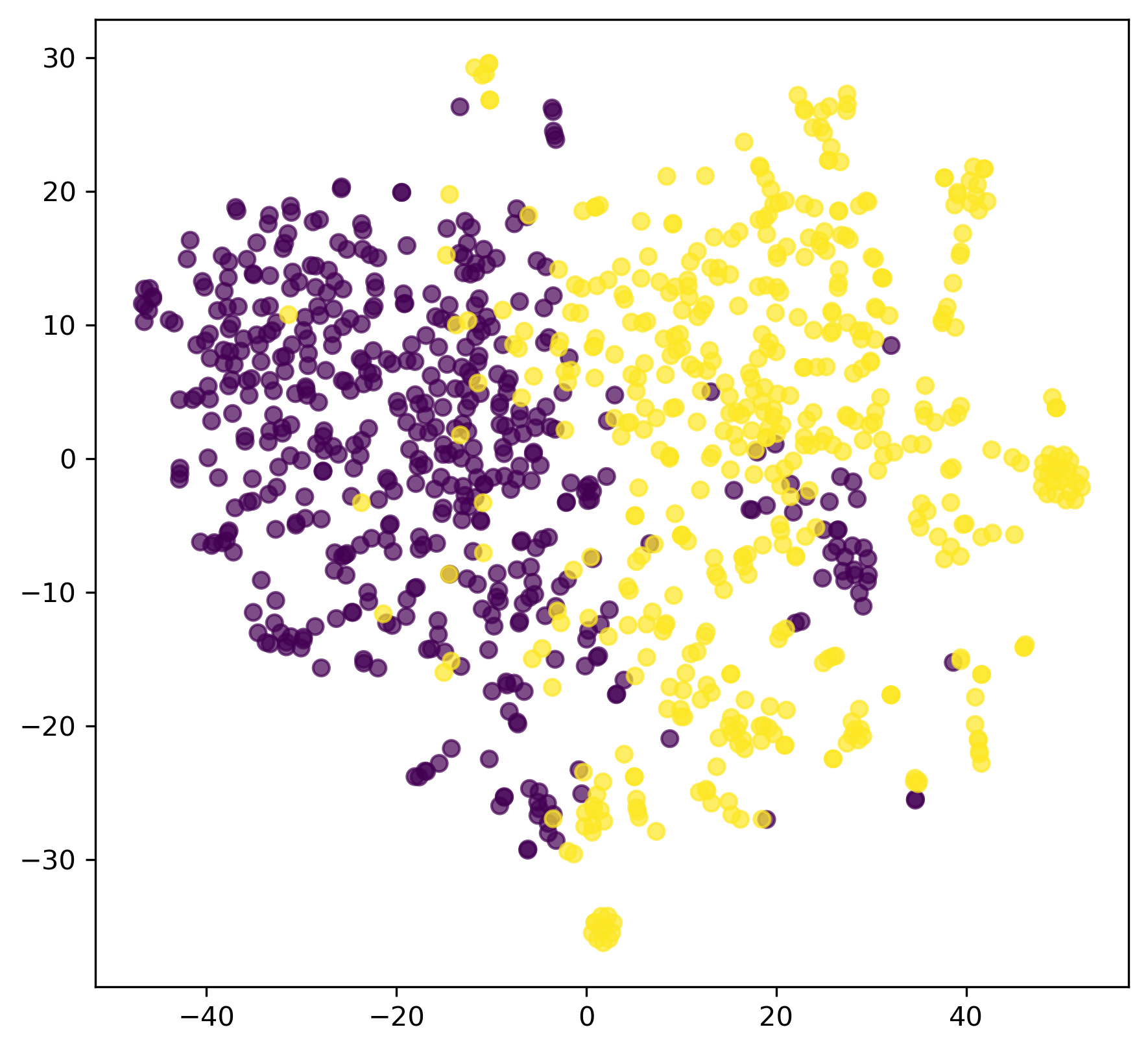}
        \caption{VAE with $\mathcal{L}_{\text{cls}}$}
        \label{fig:tsne_1}
    \end{subfigure}
    \caption{The t-SNE visualisation of latent codes coloured by class for Phishing dataset. Left: VAE without $\mathcal{L}_{\text{cls}}$ shows mixed clusters. Right: Proposed VAE with $\mathcal{L}_{\text{cls}}$ exhibits class-separable structure, explaining the lower $\delta_{\text{acc}}$.}
    \label{fig:tsne}
\end{figure}

\paragraph{Latent Space Representation}  
Figure~\ref{fig:tsne} illustrates the latent space topology generated for the Phishing dataset. While the baseline VAE produces overlapping class clusters (e.g., legitimate vs phishing URLs intermixed), our model separates classes clearly, with distinct clusters corresponding to label categories. This structured separation explains the improved $\delta_{\text{acc}}$, as classifiers trained on well-separated latent codes require less complex decision boundaries. The reduced overlap also mitigates latent space biases inherent to standard VAEs, ensuring adversarial perturbations during attack generation are less skewed by spurious class correlations. 

\subsection{Results for Task 2: Adversarial Attack Effectiveness and Imperceptibility (RQ2)}

Table~\ref{tab:asr-only} and \ref{tab:adv-results-imperceptibility} present adversarial attack performance across six tabular datasets and three model architectures. Table~\ref{tab:asr-only} focuses on attack effectiveness measured by Attack Success Rate (ASR), while Table~\ref{tab:adv-results-imperceptibility} evaluates imperceptibility through Outlier Rate and In-Distribution Success Rate (IDSR). The evaluation includes traditional input-space attacks (FGSM and PGD) and three VAE-based approaches: PGD-VAE, DeltaZ for binary classification, and our proposed method. Traditional methods serve as reference baselines to demonstrate the challenges of input-space perturbations, while the primary comparison focuses on the relative effectiveness of different VAE-based latent-space strategies.

\begin{table}[!t]
    \caption{Attack Success Rate (ASR) comparison across all datasets and models. Sample sizes vary by dataset: Adult, Phishing, Electricity, and PenDigits use 500 test samples; German uses 152 samples due to smaller dataset size; Covertype uses 497 samples to ensure equal representation across all 7 classes. DeltaZ is limited to binary classification tasks only. Best results are in \textbf{bold}, and runner-up results \underline{underlined} are shown only among VAE-based methods.}
    \label{tab:asr-only}
    \centering
    \resizebox{\textwidth}{!}{%
    \begin{tabular}{@{}llrrrrrr@{}}
        \toprule
        \multirow{2}{*}{Model} & \multirow{2}{*}{Attack}
        & Adult & Phishing & German & Electricity & PenDigits & Covertype \\
         &  & ($n=500$) & ($n=500$) & ($n=152$) & ($n=500$) & ($n=500$) & ($n=497$) \\ \midrule
        MLP & FGSM
            & 95.8 & 94.6 & 98.7 & 59.6 & 77.0 & 92.4 \\
            & PGD
            & 95.8 & 96.8 & 89.5 & 62.6 & 85.4 & 95.0 \\
            \arrayrulecolor{black!35}
            \cmidrule(l){2-8}
            \arrayrulecolor{black}
            \addlinespace[2pt]
            & PGD (VAE)
            & 59.2 & 93.0 & \textbf{18.1} & 73.2 & 66.0 & 85.1 \\
            & DeltaZ (VAE)
            & \textbf{73.4} & 82.6 & 8.1 & 65.8 & -- & -- \\
            & Ours (VAE)
            & 51.0 & \textbf{98.0} & 9.4 & \textbf{80.2} & \textbf{90.8} & \textbf{88.7} \\
        \arrayrulecolor{black!35}
        \midrule
        \arrayrulecolor{black}
        \addlinespace[2pt]
        Soft DT & FGSM
            & 92.0 & 80.6 & 100.0 & 55.0 & 73.4 & 99.0 \\
            & PGD
            & 81.8 & 83.4 & 95.4 & 61.8 & 85.0 & 99.8 \\
            \arrayrulecolor{black!35}
            \cmidrule(l){2-8}
            \arrayrulecolor{black}
            \addlinespace[2pt]
            & PGD (VAE)
            & \textbf{57.6} & \textbf{92.8} & \textbf{15.7} & 80.2 & \textbf{74.4} & 89.5 \\
            & DeltaZ (VAE)
            & 53.2 & 65.8 & 6.5 & 66.2 & -- & -- \\
            & Ours (VAE)
            & 49.4 & 74.4 & 11.1 & \textbf{88.8} & 53.8 & \textbf{93.0} \\
        \arrayrulecolor{black!35}
        \midrule
        \arrayrulecolor{black}
        \addlinespace[2pt]
        Tab Tx. & FGSM
            & 80.2 & 99.8 & 90.5 & 54.6 & 74.0 & 91.8 \\
            & PGD
            & 76.6 & 99.8 & 87.8 & 61.8 & 73.2 & 90.5 \\
            \arrayrulecolor{black!35}
            \cmidrule(l){2-8}
            \arrayrulecolor{black}
            \addlinespace[2pt]
            & PGD (VAE)
            & \textbf{42.8} & 90.2 & \textbf{18.1} & 71.6 & 69.6 & 89.1 \\
            & DeltaZ (VAE)
            & 3.8 & 85.8 & 10.7 & 62.2 & -- & -- \\
            & Ours (VAE)
            & 29.0 & \textbf{97.0} & 13.4 & \textbf{82.8} & \textbf{97.4} & \textbf{99.0} \\
        \bottomrule
    \end{tabular}
    }%
\end{table}

\begin{table}[!t]
    \caption{Adversarial attack imperceptibility performance across all datasets and models. Sample sizes vary by dataset: Adult, Phishing, Electricity, and PenDigits use 500 test samples; German uses 152 samples due to smaller dataset size; Covertype uses 497 samples to ensure equal representation across all 7 classes. Metrics include In-Distribution Success Rate (IDSR = ASR $\times$ (1 - $\mathcal{O}_r$)) and Outlier Rate ($\mathcal{O}_r$). DeltaZ is limited to binary classification tasks only. Best results are in \textbf{bold}, and runner-up results \underline{underlined} are shown only among VAE-based methods.}
    \label{tab:adv-results-imperceptibility}
    \centering
    \resizebox{\textwidth}{!}{%
    \begin{tabular}{@{}llrrrrrrrrrrrr@{}}
        \toprule
        \multirow{2}{*}{Model} & \multirow{2}{*}{Attack}
        & \multicolumn{2}{c}{Adult ($n$=500)}
        & \multicolumn{2}{c}{Phishing ($n$=500)}
        & \multicolumn{2}{c}{German ($n$=152)}
        & \multicolumn{2}{c}{Electricity ($n$=500)}
        & \multicolumn{2}{c}{PenDigits ($n$=500)}
        & \multicolumn{2}{c}{Covertype ($n$=497)} \\
        \cmidrule(r){3-4}\cmidrule(r){5-6}\cmidrule(r){7-8}\cmidrule(r){9-10}\cmidrule(r){11-12}\cmidrule(r){13-14}
         &  & IDSR  & $\mathcal{O}_r$ 
            & \ \ \ IDSR  & $\mathcal{O}_r$ 
            & \ \ IDSR  & $\mathcal{O}_r$ 
            & \ \ \ IDSR  & $\mathcal{O}_r$ 
            & \ \ \ IDSR  & $\mathcal{O}_r$ 
            & \ \ \ IDSR  & $\mathcal{O}_r$ \\ 
            \midrule
        MLP & FGSM
            & 20.8 & 78.2
            & 0.0 & 100.0
            & 21.7 & 78.0
            & 59.6 & 0.0
            & 77.0 & 0.0
            & 0.0 & 100.0 \\
            & PGD
            & 20.8 & 78.2
            & 0.0 & 100.0
            & 70.4 & 21.3
            & 62.6 & 0.0
            & 85.0 & 0.5
            & 0.0 & 100.0 \\
            \arrayrulecolor{black!35}
            \cmidrule(l){2-14}
            \arrayrulecolor{black}
            \addlinespace[2pt]
            & PGD (VAE)
            & 12.2 & 79.4
            & 21.0 & 77.4
            & \textbf{16.8} & 7.4
            & \underline{66.4} & 9.3
            & 61.8 & 6.4
            & 45.7 & 46.3 \\
            & DeltaZ (VAE)
            & \textbf{50.0} & \underline{31.9}
            & \underline{79.6} & \textbf{3.6}
            & 8.1 & \textbf{0.0}
            & 64.6 & \textbf{1.8}
            & -- & --
            & -- & -- \\
            & Ours (VAE)
            & \underline{42.6} & \textbf{16.5}
            & \textbf{80.6} & \underline{17.8}
            & \underline{9.4} & \textbf{0.0}
            & \textbf{77.8} & \underline{3.0}
            & \textbf{85.8} & \textbf{5.5}
            & \textbf{56.7} & \textbf{36.1} \\
        \arrayrulecolor{black!35}
        \midrule
        \arrayrulecolor{black}
        \addlinespace[2pt]
        Soft DT & FGSM
            & 9.4 & 89.8
            & 0.0 & 100.0
            & 11.9 & 88.1
            & 55.0 & 0.0
            & 73.2 & 0.3
            & 0.0 & 100.0 \\
            & PGD
            & 17.4 & 78.7
            & 0.0 & 100.0
            & 32.5 & 66.0
            & 61.8 & 0.0
            & 84.8 & 0.2
            & 0.0 & 100.0 \\
            \arrayrulecolor{black!35}
            \cmidrule(l){2-14}
            \arrayrulecolor{black}
            \addlinespace[2pt]
            & PGD (VAE)
            & 15.2 & 73.6
            & 25.0 & 73.1
            & \textbf{14.4} & 8.3
            & 73.4 & 8.5
            & \textbf{70.6} & 5.1
            & 52.1 & 41.8 \\
            & DeltaZ (VAE)
            & \textbf{47.8} & \textbf{10.2}
            & 64.4 & \textbf{2.1}
            & 6.5 & \textbf{0.0}
            & \underline{64.8} & \textbf{2.1}
            & -- & --
            & -- & -- \\
            & Ours (VAE)
            & \underline{43.8} & \underline{11.3}
            & \textbf{68.4} & \underline{8.1}
            & \underline{11.1} & \textbf{0.0}
            & \textbf{84.8} & \underline{4.5}
            & 51.4 & \textbf{4.5}
            & \textbf{62.4} & \textbf{32.9} \\
        \arrayrulecolor{black!35}
        \midrule
        \arrayrulecolor{black}
        \addlinespace[2pt]
        Tab Tx. & FGSM
            & 8.0 & 90.0
            & 0.0 & 100.0
            & 67.3 & 25.6
            & 54.6 & 0.0
            & 73.8 & 0.03
            & 0.0 & 1.00 \\
            & PGD
            & 13.4 & 82.5
            & 0.0 & 100.0
            & 69.4 & 20.9
            & 61.8 & 0.0
            & 73.0 & 0.03
            & 6.8 & 92.4 \\
            \arrayrulecolor{black!35}
            \cmidrule(l){2-14}
            \arrayrulecolor{black}
            \addlinespace[2pt]
            & PGD (VAE)
            & \underline{21.0} & 50.9
            & 32.0 & 64.5
            & \textbf{14.8} & 18.5
            & 69.4 & 3.1
            & 65.6 & \textbf{5.7}
            & 56.7 & 36.3 \\
            & DeltaZ (VAE)
            & 3.8 & \textbf{0.0}
            & \textbf{82.0} & \textbf{4.4}
            & 10.1 & 6.3
            & 61.6 & \textbf{1.0}
            & -- & --
            & -- & -- \\
            & Ours (VAE)
            & \textbf{26.4} & \underline{9.0}
            & \underline{80.6} & \underline{16.9}
            & \underline{12.8} & \textbf{5.0}
            & \textbf{81.0} & \underline{2.2}
            & \textbf{89.0} & 8.6
            & \textbf{72.6} & \textbf{26.6} \\
        \bottomrule
    \end{tabular}
    }%
\end{table}

\paragraph{Traditional vs VAE-based}  

Traditional attacks (FGSM and PGD) consistently achieve high ASR ranging from 54.6\% to 100\% across all datasets and models (Table~\ref{tab:asr-only}), demonstrating reliable effectiveness regardless of data characteristics. VAE-based methods show more variable performance that depends critically on VAE reconstruction quality. On datasets for which good reconstruction quality is possible, VAE methods achieve competitive effectiveness, but performance degrades significantly on poorly reconstructed datasets like the German dataset, where all VAE approaches struggle with ASR below 18.1\% while traditional methods maintain 54.6--100\% effectiveness. This dataset dependency highlights that VAE-based attacks are inherently constrained by the quality of the underlying generative model, unlike traditional methods that maintain consistent performance across diverse data conditions.

\begin{figure}[!t]
    \centering
    \begin{subfigure}[t]{0.48\columnwidth}
        \centering
        \includegraphics[width=\textwidth]{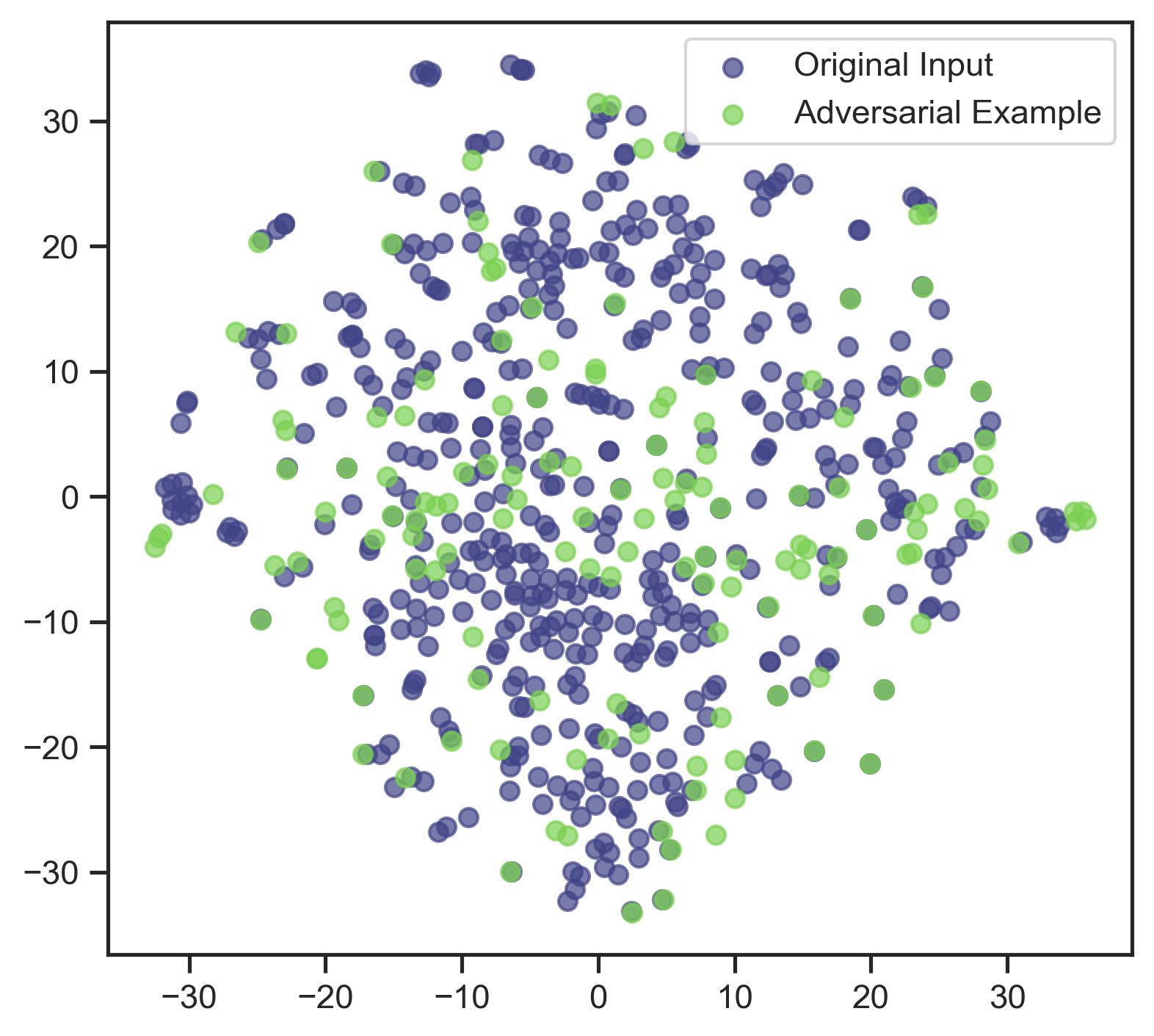}
        \caption{Phishing-MLP}
        \label{fig:md_dist_1}
    \end{subfigure}
    \begin{subfigure}[t]{0.48\columnwidth}
        \centering
        \includegraphics[width=\textwidth]{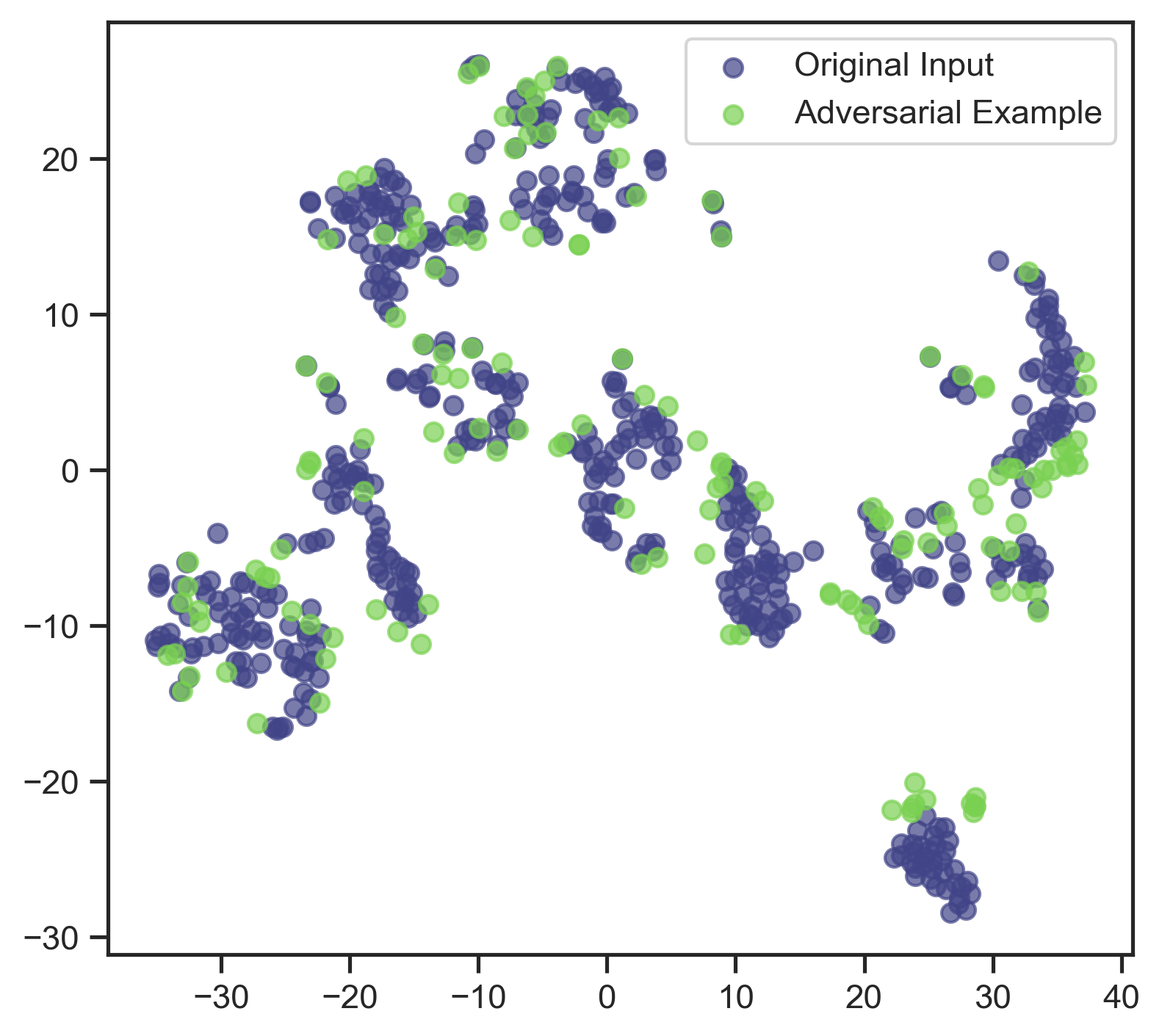}
        \caption{PenDigits-TabTransformer}
        \label{fig:md_dist_2}
    \end{subfigure}
    \caption{Latent space visualisation (t-SNE) of original inputs (blue) and adversarial examples (green) generated by VAE attack. Near-perfect overlap confirms distributional consistency, with perturbations constrained to the data manifold.}
    \label{fig:md_dist}
\end{figure}

Both input-space attacks exhibit highly variable outlier rates (Table~\ref{tab:adv-results-imperceptibility}) from 0\% to 100\% across different datasets despite using identical perturbation constraints $\epsilon=0.5$, making their imperceptibility unpredictable and dataset-dependent. VAE-based methods consistently achieve superior imperceptibility with lower and more predictable outlier rates across all datasets. The IDSR metric reveals the practical impact: traditional methods frequently achieve 0\% IDSR despite high ASR due to distributional violations, while VAE methods maintain meaningful IDSR values that closely align with their ASR. Figure~\ref{fig:md_dist} visualises how VAE-generated adversarial examples preserve distributional alignment with original data, demonstrating the fundamental advantage of latent-space perturbations for maintaining statistical consistency.

\paragraph{Performance Comparison Among VAE-based Methods}  

Among VAE-based approaches, our proposed method achieves the most consistent high ASR performance across datasets and models (Table~\ref{tab:asr-only}), with several configurations exceeding 95\%. PGD-VAE shows competitive effectiveness on well-reconstructed datasets but exhibits greater variability across different model architectures. DeltaZ demonstrates moderate ASR performance but is limited to binary classification tasks, with notably poor performance on Adult-TabTransformer (3.8\% ASR) due to TabTransformer's inherent robustness against adversarial attacks---evident in its consistently lower ASR across all methods---and DeltaZ's multiplicative perturbation design being less effective against attention-based architectures. All VAE methods struggle equally on the German dataset due to fundamental reconstruction limitations.

The IDSR analysis in Table~\ref{tab:adv-results-imperceptibility} provides the most critical evaluation metric, as it captures practical attack utility by combining effectiveness with imperceptibility. Our proposed method consistently achieves either the best or second-best IDSR performance across all dataset-model combinations, demonstrating superior reliability compared to other VAE-based approaches. While DeltaZ achieves the lowest outlier rates (0--6.3\%) on binary datasets where it applies, its limitation to binary classification severely restricts broader applicability. PGD-VAE shows the most inconsistent performance with highly variable outlier rates and correspondingly unstable IDSR values across configurations. Our proposed method demonstrates the most balanced and consistent IDSR performance across both binary and multiclass datasets through C\&W optimisation in the latent space, providing reliable performance across diverse tabular data scenarios without dataset type limitations.

\subsection{Results for Task 3: Performance Factors and Method Analysis (RQ3)}

\subsubsection{Hyperparameter Sensitivity Analysis}



\begin{figure}[htb!]
    \centering
    \begin{subfigure}[b]{0.48\columnwidth}
        \centering
        \includegraphics[width=\textwidth]{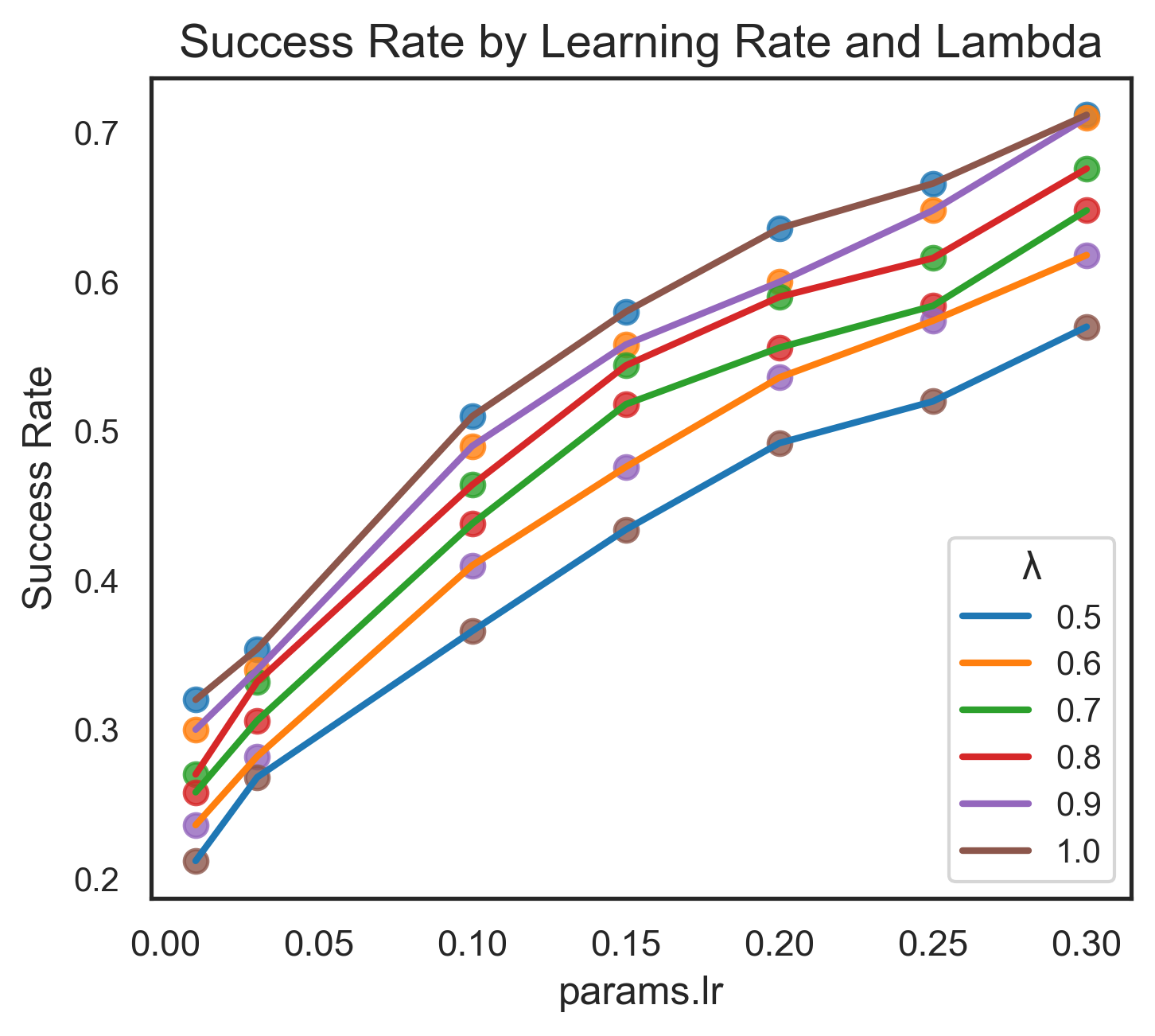}
        \caption{Success Rate vs. $\eta$ and $\lambda$.}
        \label{fig:success_rate}
    \end{subfigure}
    \hfill
    \begin{subfigure}[b]{0.48\columnwidth}
        \centering
        \includegraphics[width=\textwidth]{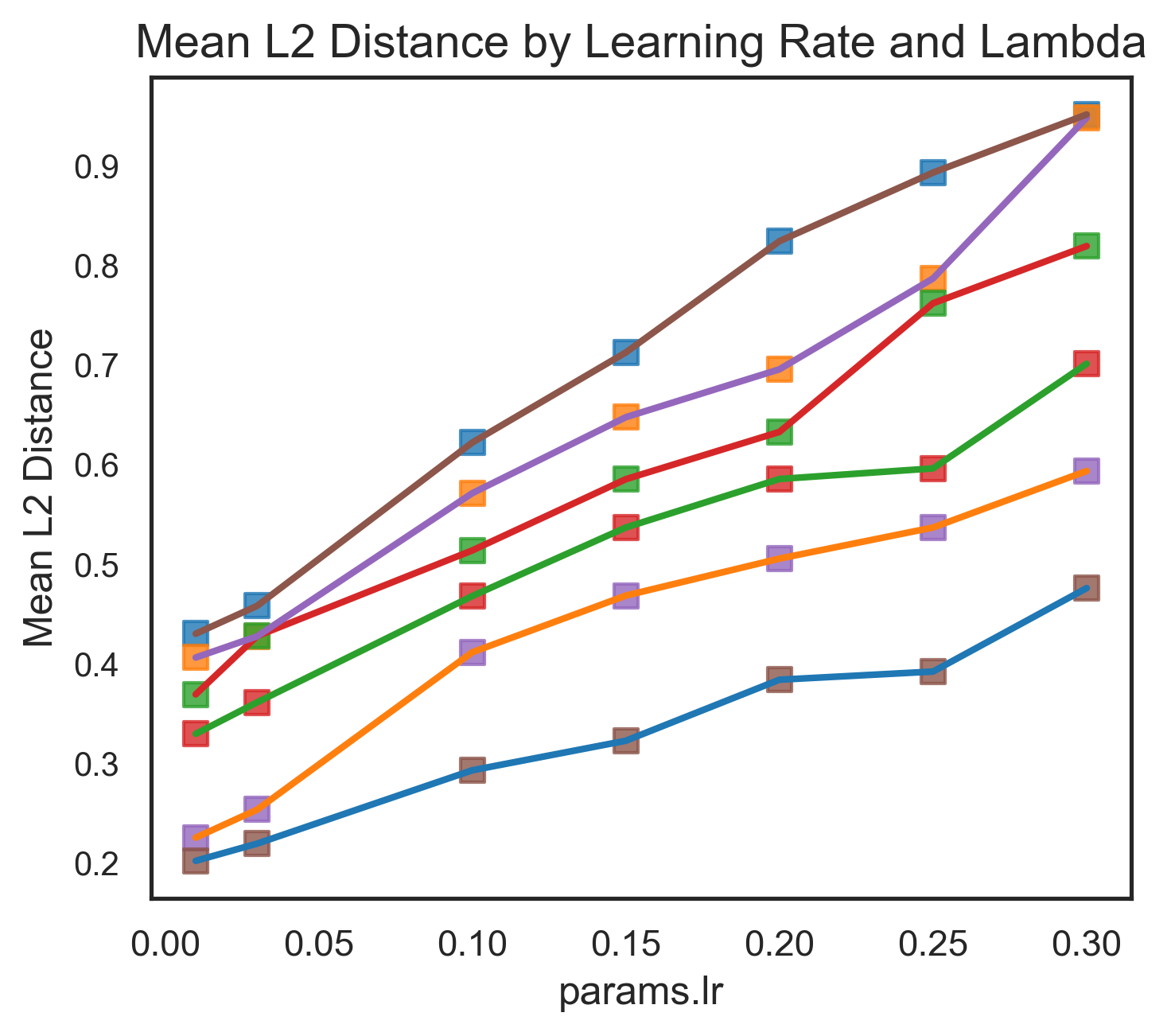}
        \caption{Mean $\ell_2$ vs. $\eta$ and $\lambda$.}
        \label{fig:l2_distance}
    \end{subfigure}
    \caption{Success rate and perturbation magnitude as functions of learning rate ($\eta$) and $\lambda$ on the Adult dataset (MLP model). Higher $\lambda$ and $\eta$ increase success rates (Figure~\ref{fig:success_rate}) but also amplify perturbation magnitudes (Figure~\ref{fig:l2_distance}), with diminishing returns beyond $\eta > 0.2$. Optimal configurations balance mid-range $\lambda$ (e.g., $\lambda = 0.7$--$0.9$) with $\eta = 0.1$--$0.2$.}
    \label{fig:side_by_side}
\end{figure}

Figures~\ref{fig:success_rate} and~\ref{fig:l2_distance} reveal the distinct roles of $\lambda$ (regularisation weight) and $\eta$ (learning rate) in adversarial attack performance, demonstrating clear trade-offs between effectiveness and imperceptibility.

\paragraph{The Role of \(\lambda\)}  
The regularisation weight \(\lambda\) governs the trade-off between attack success rate (ASR) and imperceptibility. Higher \(\lambda\) values prioritise maximising ASR over minimising \(\ell_2\) perturbations, achieving stronger misclassification at the cost of larger, more detectable perturbations. As depicted in Fig.~\ref{fig:success_rate}, \(\lambda = 1.0\) yields a 67\% ASR with \(\ell_2 = 0.85\), whereas \(\lambda = 0.5\) reduces these to 52\% and \(\ell_2 = 0.45\). This behaviour stems from the C\&W objective: \(\lambda\) scales the margin-based loss term, encouraging misclassification even if perturbations breach imperceptibility thresholds.  

\paragraph{The Role of \(\eta\)}  
The learning rate \(\eta\) controls optimisation stability under the fixed iteration budget (\(T = 500\)). Larger \(\eta\) (e.g., \(\eta = 0.2\)) accelerates convergence, enabling the optimiser to find effective perturbations within fewer iterations. However, this risks overshooting optimal solutions, leading to erratic \(\ell_2\) growth (Figure~\ref{fig:l2_distance}) and undermining imperceptibility. Smaller values of \(\eta\) (e.g., \(\eta = 0.01\)) stabilise training but may fail to converge within \(T = 300\), leaving perturbations suboptimal. For example, with \(\eta = 0.01\), gradients update too cautiously, exhausting the iteration budget before reaching the decision boundary. This highlights a critical trade-off: larger \(\eta\) priorities convergence within \(T = 500\) at the cost of instability, while a smaller \(\eta\) sacrifices convergence speed for smoother optimisation.  

\paragraph{Optimal Configuration}  
Overall, $\lambda$ and $\eta$ jointly determine the balance between attack strength and imperceptibility. 
While $\lambda$ directly modulates the optimisation objective by weighting misclassification pressure against latent-space distortion, $\eta$ governs the dynamic trajectory of that optimisation. 
Empirically, excessively large $\lambda$ values or aggressive $\eta$ schedules tend to push perturbations beyond the smooth regions of the latent manifold, causing noticeable distributional shifts. 
Conversely, moderate configurations (e.g., $\lambda=0.7$--$0.9$, $\eta=0.1$--$0.2$) achieve a favourable equilibrium, producing successful yet on-manifold adversarial examples.

\subsubsection{Sparsity Control Analysis}

While our proposed method achieves superior imperceptibility through on-manifold perturbations by addressing proximity (via $\ell_2$ distance), sensitivity (through Z-score normalisation), and deviation (via VAE manifold constraints), an important aspect of imperceptibility~\citep{he2025investigating} remains unaddressed---the \textit{sparsity} of feature perturbation.

\paragraph{Challenges in Latent Space Sparsity Control} 

Unlike input-space attacks where sparsity can be directly enforced through feature selection, controlling sparsity in latent space presents unique challenges. The VAE's encoder projects input features into a dense latent representation where individual dimensions do not correspond directly to input features. Consequently, perturbations in latent space $\delta$ may affect multiple input features upon decoding, complicating direct sparsity control.

Incorporating an $\ell_0$ norm constraint into our objective function (Eq.~\ref{eq:cw-obj}) to limit the number of modified features can be achieved as follows:

\begin{equation}\label{eq:cw-obj-spa}
\delta^* = \arg\min_{\delta} \bigg[ \lambda \cdot \max\Big( Z(\tilde{x})_y - \max_{i \neq y} Z(\tilde{x})_i + \kappa, 0\Big) + \|\delta\|_2 + \underbrace{\alpha \|\tilde{x} - x\|_0}_{\substack{\text{Sparsity Control}}} \bigg],
\end{equation}

\noindent
where $\alpha$ controls the sparsity penalty and $|\tilde{x} - x|_0$ counts the number of modified features in the reconstructed space. However, the non-differentiable nature of the $\ell_0$ norm renders this approach incompatible with gradient-based optimisation.

\paragraph{Differentiable Approximation of $\ell_0$}

To address the non-differentiable nature of the $\ell_0$ norm, we explore differentiable approximations. We replace $|\tilde{x} - x|_0$ in Eq.~(\ref{eq:cw-obj-spa}) with a sigmoid-based smooth approximation \citep{louizos2017learning}:
$|\tilde{x} - x|_0 \approx \sum_i \sigma(\beta |\tilde{x}_i - x_i| - \gamma),$
where $\sigma(\cdot)$ is the sigmoid function, $\beta = 20$ controls the steepness of the transition, and $\gamma = 0.1$ sets the threshold for considering a feature as ``perturbed''. We evaluate this approximation across different sparsity penalty weights $\alpha \in \{0.01, 0.1, 0.5, 1\}$.

\begin{table}[htb!]
\caption{Sparsity experimental results on Adult dataset with MLP model using differentiable approximation of $\ell_0$. Different sparsity penalty weights ($\alpha$) show minimal impact on actual sparsity metrics despite theoretical expectations.}
\label{tab:sparsity_results}
\centering
\begin{tabular}{@{}crrr@{}}
\toprule
\textbf{$\alpha$} & \textbf{ASR (\%)} & \textbf{$\ell_0$ Norm} & \textbf{Sparsity Rate (\%)} \\ \midrule
baseline & 51.00 & 4.66 & 38.86 \\
0.01 & 51.20 & 4.65 & 38.74 \\
0.1 & 52.80 & 4.66 & 38.83 \\
0.5 & 55.40 & 4.82 & 40.19 \\
1.0 & 54.00 & 4.88 & 40.65 \\
\bottomrule
\end{tabular}
\end{table}

Our experiments on the Adult dataset with MLP reveal that the differentiable $\ell_0$ approximation fails to achieve meaningful sparsity control. As shown in Table~\ref{tab:sparsity_results}, varying the sparsity weight $\alpha$ across different values shows minimal impact on actual sparsity metrics. The average number of changed features remains consistently around 4.6-4.9 across all $\alpha$ values, with sparsity percentages hovering near 38-40\%. This indicates that the sigmoid approximation, while theoretically sound, did not produce the expected sparsity improvements in our experimental setting. 

The limited effectiveness of the differentiable $\ell_0$ approximation can be attributed to two factors. First, the smooth sigmoid relaxation weakens the discrete sparsity constraint by penalising all feature deviations proportionally, rather than enforcing exact zeros. As a result, small but non-zero perturbations are still favoured by gradient descent. Second, the latent representation learned by the VAE entangles feature dependencies, meaning that small perturbations in latent space can simultaneously affect multiple reconstructed features. These correlated activations dilute the influence of the sparsity penalty in the reconstructed space, preventing the optimiser from isolating specific features for modification.

\paragraph{Greedy Search-based Feature Selection}

Beyond differentiable approximations, we also developed a two-stage approach to achieve sparsity. First, we perform the standard VAE attack optimisation in latent space to generate an initial adversarial example using the existing misclassification and latent space perturbation losses. Second, we analyse the generated adversarial example by ranking features according to their contribution to misclassification and iteratively removing or reducing feature modifications while preserving attack effectiveness. However, our experiments across all datasets and models showed that the greedy approach achieved minimal success, reducing the number of modified features in only one case (Phishing-MLP), indicating that the greedy search method failed to identify effective sparse perturbations. This indicates that most features modified by the VAE attack are essential for maintaining adversarial effectiveness.

\paragraph{Convex Relaxation via $\ell_1$ Norm} We investigated $\ell_1$ norm~\citep{tibshirani1996regression} as a convex relaxation of the non-differentiable $\ell_0$ constraint by replacing the sparsity term in Eq.~\ref{eq:cw-obj-spa} with $\alpha |\tilde{x} - x|_1$. As shown in our experimental results (Table~\ref{tab:l1_results}), the $\ell_1$ penalty demonstrates a clear trade-off between attack effectiveness and sparsity control. Higher values of $\alpha$ (e.g., $\alpha = 1.0$) significantly reduce the $\ell_1$ norm from 4.45 to 2.55 and improve sparsity rates marginally (38.86\% to 37.93\%), but at the cost of substantially decreased attack success rates (51.00\% to 29.00\%). This trade-off reflects the fundamental limitation of $\ell_1$ regularisation: while it encourages smaller perturbations across all features, it does not enforce the discrete constraint of exactly zero modifications that true $\ell_0$ sparsity requires.

\begin{table}[htb!]
\caption{$\ell_1$ regularisation results on Adult dataset with MLP model. Higher $\alpha$ values improve sparsity control but significantly reduce attack success rates.}
\label{tab:l1_results}
\centering
\begin{tabular}{@{}crrrr@{}}
\toprule
\textbf{$\alpha$} & \textbf{ASR (\%)} & \textbf{$\ell_0$ Norm} & \textbf{$\ell_1$ Norm} & \textbf{Sparsity Rate (\%)} \\ \midrule
baseline & 51.00 & 4.66 & 4.45 & 38.86 \\
0.01 & 50.60 & 4.66 & 4.44 & 38.80 \\
0.1 & 45.80 & 4.59 & 4.27 & 38.21 \\
0.25 & 40.00 & 4.59 & 3.86 & 38.25 \\
0.5 & 34.80 & 4.53 & 3.53 & 37.79 \\
0.75 & 30.40 & 4.44 & 3.03 & 37.01 \\
1.0 & 29.00 & 4.55 & 2.55 & 37.93 \\
\bottomrule
\end{tabular}
\end{table}

The $\ell_1$ penalty exhibited similar behaviour for related reasons. Although $\ell_1$ regularisation promotes small overall perturbations, it distributes shrinkage uniformly across features rather than driving individual coefficients exactly to zero. In the latent-to-input mapping of the VAE, this uniform shrinkage translates into small correlated adjustments to many features, leading to modest improvements in sparsity but substantial degradation in attack success.

\paragraph{Analysis of Sparsity Relaxations}
Overall, these results indicate that continuous relaxations of sparsity (either sigmoid-based or $\ell_1$-based) are poorly aligned with the discrete and entangled nature of latent-space perturbations. Our exploration of explicit sparsity constraints reveals that our VAE-based attack framework already possesses inherent sparsity control mechanisms. The VAE's learned latent space structure enables small, targeted perturbations to translate into changes affecting only the most relevant features, as semantically related features are encoded together in the compact representation. Additionally, the $\ell_2$ norm penalty in latent space naturally encourages minimal perturbations, and since the VAE maps similar inputs to nearby latent points, this inherently leads to modifying only the features most critical for classification. While explicit sparsity methods like differentiable $\ell_0$ approximations and $\ell_1$ regularisation showed limited effectiveness, these findings suggest that the VAE framework's implicit sparsity control through learned representations may be more effective than direct constraint enforcement for tabular adversarial generation.

\paragraph{Scope and Future Considerations}
Our sparsity control experiments primarily evaluate quantitative properties—such as the number of modified features and the magnitude of perturbations—without incorporating semantic or domain-specific constraints. This design choice ensures that the proposed framework remains general and applicable to diverse tabular datasets without relying on domain priors. Semantic imperceptibility aspects, including feature immutability and feasibility, have been extensively discussed in our earlier work on imperceptibility properties for tabular data~\citep{he2025investigating}. The present study focuses on the quantitative subset of these properties to maintain methodological generality. Incorporating feature-importance weighting or domain-aware feasibility checks into sparsity control mechanisms represents a promising direction for future work.

\subsubsection{Generative Model Comparison: VAE vs GAN}

The choice of generative model significantly impacts the quality and imperceptibility of adversarial examples in tabular domains. While our approach employs VAEs for latent space perturbations, GANs represent a popular alternative for generative modelling. To validate our architectural choice and demonstrate the specific advantages of VAEs for tabular adversarial generation, we conducted a comparative evaluation of reconstruction quality.

\begin{table}[ht!]
\caption{Reconstruction performance comparison between VAE and GAN approaches across three datasets. Metrics include accuracy delta ($\delta_{\text{acc}}$), accuracy retention ratio ($A_\text{ret}$), MSE, coefficient of determination ($R^2$), cosine similarity ($\cos$), Pearson correlation ($\rho$), and categorical reconstruction accuracy ($x^{\text{cat}}_{\text{acc}}$). Best results in \textbf{bold}.}
\label{tab:gan_comparison}
\centering
\begin{tabular}{@{}lcrrrrrrr@{}}
\toprule
Dataset & Model & $\delta_{\text{acc}}$ & $A_\text{ret}$ & MSE & $R^2$ & $\cos$ & $\rho$ & $x^{\text{cat}}_{\text{acc}}$ \\ \midrule
Adult & VAE & \textbf{0.0030} & \textbf{0.9965} & \textbf{0.0308} & \textbf{0.9679} & \textbf{0.9851} & \textbf{0.9854} & \textbf{0.9864} \\
 & GAN & 0.0222 & 0.9740 & 0.1287 & 0.8659 & 0.9341 & 0.9345 & 0.1156 \\
  \addlinespace
Phishing &  VAE & 0.0184 & 0.9809 & \textbf{0.1487} & \textbf{0.8320} & \textbf{0.9124} & \textbf{0.9130} & \textbf{0.9552} \\
 & GAN & \textbf{0.0004} & \textbf{0.9996} & 0.4876 & 0.4491 & 0.6759 & 0.6764 & 0.8617 \\
 \addlinespace
PenDigits &  VAE & \textbf{0.0059} & \textbf{1.0023} & \textbf{0.0588} & \textbf{0.9417} & \textbf{0.9706} & \textbf{0.9707} & N/A \\
 & GAN & 0.0241 & 0.9754 & 0.3077 & 0.6947 & 0.8349 & 0.8349 & N/A \\
\bottomrule
\end{tabular}
\end{table}

\begin{figure}[ht!]
    \centering
    \includegraphics[trim={0 3cm 0 0},clip,width=\textwidth]{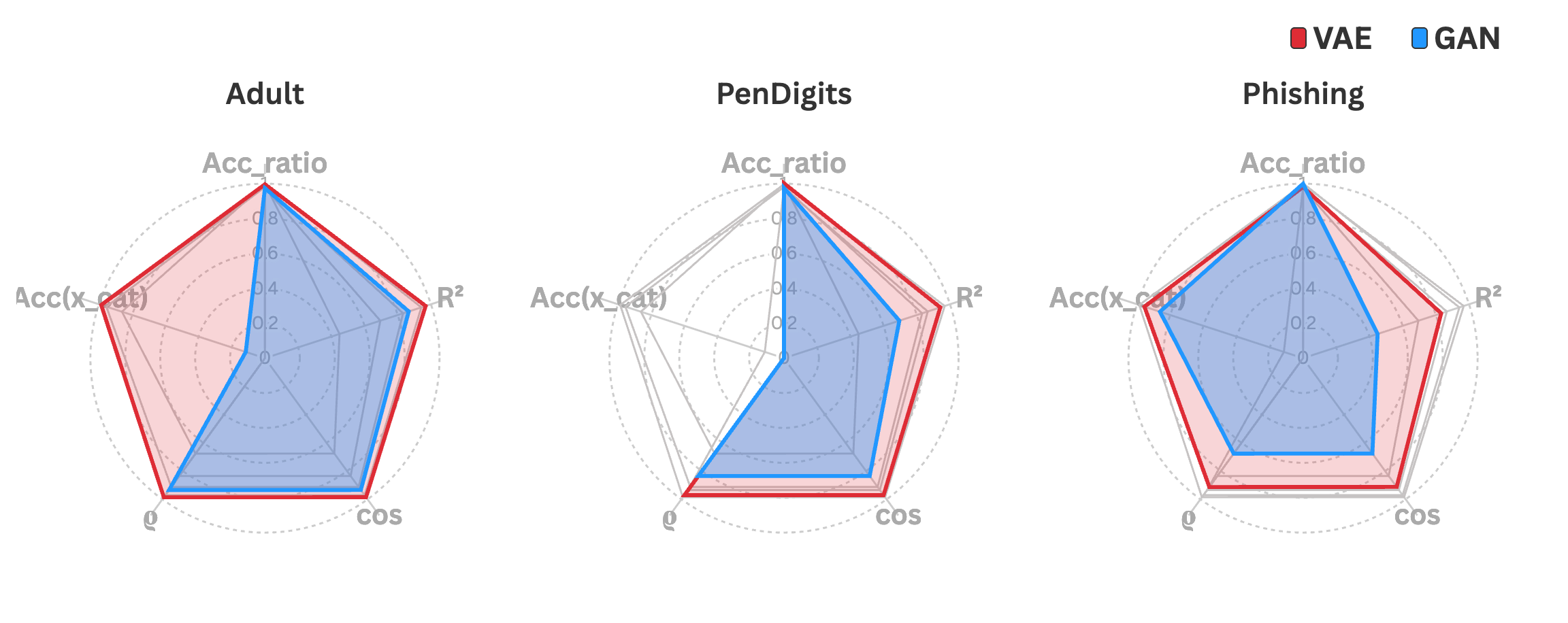}
    \caption{Radar-chart comparison of reconstruction performance between VAE and GAN models across three representative datasets. Axes correspond to accuracy retention ratio (\(A_{\text{ret}}\)), coefficient of determination (\(R^2\)), cosine similarity (\(\cos\)), Pearson correlation (\(\rho\)) and categorical reconstruction accuracy ($x^{\text{cat}}_{\text{acc}}$). Larger enclosed areas indicate better overall reconstruction quality. The \textit{PenDigits} dataset contains only numerical attributes; thus, its categorical reconstruction accuracy axis is zero.}
    \label{fig:vae_gan_radar}
\end{figure}

Table~\ref{tab:gan_comparison} compares VAE and GAN reconstruction across three datasets. The VAE consistently outperforms the GAN across most metrics, achieving substantially better numerical reconstruction with higher $R^2$ scores and lower MSE values, while maintaining superior categorical reconstruction accuracy. The difference is most apparent in categorical reconstruction, where GANs struggle significantly with discrete variables. Although GANs occasionally show competitive performance for isolated metrics, this is typically accompanied by degraded reconstruction quality across other dimensions, indicating inadequate preservation of overall data characteristics. To complement the tabulated results, Figure~\ref{fig:vae_gan_radar} visualises the comparative performance across five key reconstruction metrics (\(A_{\text{ret}}\), \(R^2\), \(\cos\), \(\rho\), $x^{\text{cat}}_{\text{acc}}$). The VAE encloses consistently larger areas across all datasets, reflecting more stable reconstruction quality and stronger alignment between numerical and categorical feature spaces compared with the GAN baseline.

\paragraph{Scope of GAN comparison}
The GAN baseline in Table~\ref{tab:gan_comparison} was evaluated only on reconstruction quality because conventional tabular GANs do not provide an explicit inference mapping \(x\mapsto z\). A fair latent-space adversarial comparison requires an encoder or inference mechanism (for example BiGAN/ALI \citep{donahue2016adversarial} or a separately trained encoder) that enables mapping real inputs into the generator latent space. Implementing such encoder-enabled GAN variants or adversarially learned inference introduces additional architectural complexity and hyperparameter choices that are not directly comparable to the VAE setup used throughout this work; therefore, evaluating GAN-based adversarial generation is left for future work. We emphasise that reconstruction quality, especially for categorical features, is an essential prerequisite for on-manifold adversarial generation, and the reconstruction results reported here motivate our choice of VAEs for the main experiments.

These reconstruction differences reflect fundamental architectural characteristics that directly affect adversarial generation. GANs optimise a min-max objective without explicit reconstruction constraints~\citep{goodfellow2014generative}, providing no incentive to preserve exact feature values or categorical mappings, and they often suffer from mode collapse where the generator fails to capture full data diversity~\citep{arjovsky2017towards}. In tabular settings, this instability and lack of invertibility make it difficult to control latent perturbations or ensure that generated samples remain consistent with valid feature combinations. In contrast, VAEs employ explicit reconstruction loss within the evidence lower bound (ELBO) framework, ensuring that the learned latent space preserves semantic correspondence between categorical and numerical features and remains continuous and well-structured. Because the KL divergence term regularises the posterior $q_\phi(z|x)$ toward the prior $p(z)$, decoded samples lie within high-probability regions of the data distribution, providing a probabilistic guarantee that generated examples stay on-manifold. Overall, VAEs offer greater stability, interpretability, and controllability, making them more suitable for generating imperceptible adversarial examples on tabular data.

\section{Conclusion}


This study introduces a comprehensive VAE-based framework for generating imperceptible adversarial attacks on tabular data, addressing the unique challenges posed by mixed categorical and numerical features. Through systematic evaluation across six diverse datasets and three model architectures, we demonstrate that latent space perturbations offer fundamental advantages over traditional input-space approaches for maintaining statistical consistency with original data distributions.

Our key contributions include the development of a mixed-input VAE architecture that unifies heterogeneous tabular features into a coherent latent representation, and the introduction of the In-Distribution Success Rate (IDSR) metric which captures the practical utility of adversarial attacks by combining effectiveness with imperceptibility. The experimental results reveal that while traditional attacks achieve high attack success rates, they suffer from unpredictable and severe distributional violations across datasets. Compared to other VAE-based approaches adapted from image domain methods, our method demonstrates superior balance between attack effectiveness and imperceptibility, providing the most reliable performance across diverse tabular data scenarios.


The comprehensive analysis of performance factors reveals important insights for practical deployment. VAE-based attacks are inherently dependent on reconstruction quality, as demonstrated by the poor performance on datasets with insufficient training data. Our exploration of sparsity control mechanisms shows that while explicit constraints like $\ell_0$ and $\ell_1$ regularisation provide limited improvements, the VAE framework possesses inherent sparsity properties through its learned latent representations. The comparison with GANs validates our architectural choice, demonstrating VAE's superior reconstruction fidelity and categorical feature preservation which are essential for generating realistic adversarial examples.

These findings underscore the importance of using manifold-aligned perturbations for creating realistic adversarial attacks on mixed-type tabular data, offering a more reliable and practical approach to adversarial example generation in this domain. Future research directions include addressing the training data sufficiency requirements for effective VAE learning on small datasets, developing more sophisticated sparsity control methods that leverage VAE's latent structure, and investigating the framework's applicability to adversarial robustness evaluation in production ML systems.

Future work can extend the proposed framework in several directions. Incorporating more advanced generative architectures, such as tabular diffusion models and normalising flows, may improve reconstruction fidelity and enrich the latent manifold, thereby increasing the realism of decoded adversarial examples. It would also be valuable to incorporate semantic constraints into the perturbation design so that generated examples respect feature-level meaning and inter-feature dependencies, improving plausibility for domain-specific applications. Finally, adaptive control of attack confidence could be explored to achieve a more flexible balance between attack strength and imperceptibility, allowing the perturbation process to adjust dynamically to model confidence or data characteristics.

\section*{Code Availability}

The implementation code, including data processing scripts and experimental pipelines, is openly available at \url{https://github.com/ZhipengHe/VAE-TabAttack}

\section*{CRediT authorship contribution statement}


\textbf{Zhipeng He:} Conceptualisation, Methodology, Software, Investigation, Writing -- Original Draft \& Revision, Visualisation. 
\textbf{Alexander Stevens:} Conceptualisation, Methodology, Investigation, Visualisation, Writing -- Review \& Editing.
\textbf{Chun Ouyang:} Conceptualisation, Methodology, Investigation, Writing -- Original Draft. 
\textbf{Johannes De Smedt:} Methodology, Investigation, Writing -- Review \& Editing. 
\textbf{Alistair Barros:} Supervision. 
\textbf{Catarina Moreira:} Methodology, Investigation, Supervision.

\section*{Acknowledgements}

The reported research forms part of part of a Ph.D. project supported by QUT Postgraduate Research Award (QUTPRA) at Queensland University of Technology (QUT), Australia. This work was supported in part by the Research Foundation Flanders (FWO) under grant number V440623N as well as grant number G039923N, by KU Leuven, Belgium under project 3H200414, and Internal Funds KU Leuven under grant number C14/23/031. The work reported in this article was also partially supported under the auspices of the UNESCO Chair on AI \& VR.

\appendix



\bibliographystyle{elsarticle-num-names} 
\bibliography{bibliography}



\end{document}